\documentclass[runningheads]{llncs}

\usepackage[utf8]{inputenc} 
\usepackage[T1]{fontenc}    
\usepackage{url}            
\usepackage{booktabs}       
\usepackage{amsfonts}       
\usepackage{nicefrac}       
\usepackage{microtype}      
\usepackage{multirow}
\usepackage{enumitem}
\usepackage{amsmath,amsfonts}
\usepackage{bm}
\usepackage{graphicx}
\usepackage{caption}
\usepackage{listings}
\usepackage{textcomp}
\usepackage{xcolor}
\usepackage{colortbl}
\usepackage{subcaption}
\usepackage{tcolorbox}
\usepackage{xspace}
\usepackage{flushend}
\usepackage{pifont}

\usepackage[ruled,linesnumbered]{algorithm2e}    
\usepackage{algpseudocode} 

\newcommand*{\method}{{\emph{ArchRepair}}}

\def\argmin{\operatornamewithlimits{arg\,min}}
\def\argmax{\operatornamewithlimits{arg\,max}}

\newcommand{\figref}[1]{Fig.~\ref{#1}}
\newcommand{\reqref}[1]{Eq.~\eqref{#1}}
\newcommand{\secref}[1]{Sec.~\ref{#1}}
\newcommand{\tableref}[1]{Table~\ref{#1}}

\definecolor{dg}{rgb}{0,0.694,0.298}
\definecolor{purple}{rgb}{0.4,0.176,0.569}
\definecolor{royalblue}{RGB}{65,105,225}
\definecolor{colorsteps}{rgb}{0.83, 0.83, 0.83}
\definecolor{tab_red}{rgb}{1,0.76,0.71}

\usepackage{hyperref}
\hypersetup{
    colorlinks=true,
    linkcolor=red,
    citecolor=royalblue,
    filecolor=magenta,      
    urlcolor=magenta,
}

\makeatletter
\DeclareRobustCommand\onedot{\futurelet\@let@token\@onedot}
\def\@onedot{\ifx\@let@token.\else.\null\fi\xspace}
\def\eg{\emph{e.g}\onedot} 
\def\ie{\emph{i.e}\onedot} 
 
\def\etc{\emph{etc}\onedot} 
\def\wrt{w.r.t\onedot} 
\def\etal{\emph{et al}\onedot}
\makeatother

\newcommand{\repeatthanks}{\textsuperscript{\thefootnote}}

\begin{document}

\title{\emph{ArchRepair}: Block-Level Architecture-Oriented Repairing for Deep Neural Networks}
\titlerunning{\emph{ArchRepair}: Block-Level Architecture-Oriented Repairing for DNNs}
%

\author{Hua Qi\inst{1}\thanks{Both authors contributed equally to this research.}\and
Zhijie Wang\inst{2}\repeatthanks\and 
Qing Guo\inst{3}\thanks{Corresponding author} \and
Jianlang Chen\inst{1} \and\\
Felix Juefei-Xu\inst{4} \and 
Lei Ma\inst{1,5,2} \and
Jianjun Zhao\inst{1}}

\authorrunning{H. Qi, Z. Wang, Q. Guo, J. Chen, F. Juefei-Xu, L. Ma, and J. Zhao}
%
\institute{Kyushu University, Japan\and
University of Alberta, Canada\and 
Nanyang Technological University, Singapore\and
Alibaba Group, USA\and
Alberta Machine Intelligence Institute (Amii), Canada
}

\maketitle              

\begin{abstract}
Over the past few years, deep neural networks (DNNs) have achieved tremendous success and have been continuously applied in many application domains. However, during the practical deployment in the industrial tasks, DNNs are found to be erroneous-prone due to various reasons such as overfitting, lacking robustness to real-world corruptions during practical usage. To address these challenges, many recent attempts have been made to repair DNNs for version updates under practical operational contexts by updating weights (\ie, network parameters) through retraining, fine-tuning, or direct weight fixing at a neural level.
Nevertheless, existing solutions often neglect the effects of neural network architecture and weight relationships across neurons and layers. In this work, as the first attempt, we initiate to repair DNNs by jointly optimizing the architecture and weights at a higher (\ie, block) level.

We first perform empirical studies to investigate the limitation of whole network-level and layer-level repairing, which motivates us to explore a novel repairing direction for DNN repair at the block level. To this end, we need to further consider techniques to address two key technical challenges, \ie, \textit{block localization}, where we should localize the targeted block that we need to fix; and how to perform \textit{joint architecture and weight repairing}.  Specifically, we first propose \textit{adversarial-aware spectrum analysis for vulnerable block localization} that considers the neurons' status and weights' gradients in blocks during the forward and backward processes, which enables more accurate candidate block localization for repairing even under a few examples. Then, we further propose the \textit{architecture-oriented search-based repairing} that relaxes the targeted block to a continuous repairing search space at higher deep feature levels. By jointly optimizing the architecture and weights in that space, we can identify a much better block architecture. We implement our proposed repairing techniques as a tool, named \method{}, and conduct extensive experiments to validate the proposed method. The results show that our method can not only repair but also enhance accuracy \& robustness, outperforming the state-of-the-art DNN repair techniques.
\end{abstract}

\keywords{Deep Learning\and DNN Repair\and Network Architecture Search}

\section{Introduction}\label{sec:introduction}

Modern high-capacity deep neural networks (DNNs) have shown astounding performance in many automated computer vision tasks ranging from complex scene understanding for autonomous driving \cite{Wang_2021_CVPR,Choi_2021_CVPR,Chen_2021_CVPR,Prakash_2021_CVPR,Luo_2021_CVPR,Li_2021_ICCV}, to accurate DeepFake media detection \cite{juefei2021countering,dolhansky2020deepfake}; from challenging medical imagery grading and diagnosis \cite{yim2020predicting,fan2020inf,cheng2020adversarial,icme21_xray}, to billion-scale consumer applications such as the face authentication for mobile payment, \etc. Many of the tasks are safety- and mission-critical and the reliability of the deployed DNNs are of utmost importance. However, over the years, we have come to realize that the existence of unintentional (natural degradation corruptions) and intentional (adversarial perturbations) examples such as \cite{tmm21_pasadena,iccv21_advmot,arxiv21_ara,ijcai21_ava,arxiv21_advhaze,neurips20_abba,eccv20_spark,zhai2020s_rain,acmmm20_amora,iccv21_flat,icme21_xray,gao2020making,arxiv21_advbokeh,cheng2020adversarial} is a stark reminder that DNNs are vulnerable.

To tackle the DNN's vulnerability issues, many researchers have resorted to DNN repairing that aims at fixing the faulty DNN weights with the guidance of some specific repairing optimization criteria. An analogy to this is the traditional software repairing in the software engineering literature \cite{gazzola2017automatic}. However, general-purpose DNN repairing may not always be feasible in practice, due to (1) the difficulty of generalizing DNNs to any arbitrary unseen scenarios, and (2) the difficulty of generalizing DNNs to seen scenarios but with unpredictable, volatile, and ever-changing deployed environment.  
For these reasons, a more practical DNN repairing strategy is to work under some assumptions of practical contexts and to perform task-specific and environment-aware DNN repairing where the model gap is closed up for a certain scenario/environment, or a set of scenarios/environments.

Compared to existing DNN repair work (\eg, \cite{ma2018mode,zhang2019apricot,sohn2019arachne,gao2020sensei,ren2020fewshot,yu2021deeprepair}), this work takes the DNN repairing to a whole new level, quite literally, where we are performing \textbf{block-level} architecture-oriented repairing as opposed to network-level, layer-level, and neuron-level repairing.
As we will show in the following sections that block-level repairing, being a midpoint sweet spot in terms of network module granularity, offers a good trade-off between network accuracy and time consumption for that just repairing some specific weights in a layer neglects the relationship between different layers while repairing the whole network weights leads to high cost. In addition, block-level repairing allows us to locally adjust not only the weights but also the network architecture within the block very effectively and efficiently. 

To this end, as the first attempt, we repair DNNs by jointly optimizing the architecture and weights at the block level in this work. 
The modern block structure stems from the philosophy of VGG nets \cite{Simonyan2015ICLR} and is generalized to a common designing strategy in the state-of-the-art architectures \cite{he2016resnet} (\eg, ResNet) and optimization method \cite{liu2018darts}.
To validate its importance for block-level repairing, we first study the drawbacks of network-level and layer-level repairing, which motivates us to explore a novel research granularity and repairing direction.
Eventually, we identified that block-level architecture-oriented DNN repair is a promising direction. In order to achieve this, we need to address two challenges, \ie, \textit{block localization} and \textit{joint architecture and weight repairing}.
For the first challenge, we propose the \textit{adversarial-aware spectrum analysis for vulnerable block localization} that considers the neuron suspiciousness and weights' gradients in blocks during the forward and backward processes when evaluating a series of examples. This method enables more precise block localization even under few-shot examples.
In terms of the second challenge, we propose the \textit{architecture-oriented search-based repairing} that relaxes the targeted block to a continuous search space. The space consists of several nodes and edges where the node represents deep features and the edge is an operation to connect two nodes. 
By jointly optimizing the architecture and weights in that space, our method is able to find a much better block architecture for a specific repairing target.
We conduct extensive experiments to validate the proposed repairing method and find that our method can not only enhance the accuracy but also the robustness across various corruptions. 
The different DNN models repaired with our technique perform better than the original one on both clean and corrupted data, with an average 3.939\% improvement on clean data and 7.79\% improvement on corrupted data, establishing vigorous general repairing capability on most of the DNN architectures.

Overall, the contribution of this paper is summarized as follows:
\begin{itemize} 
    \item We propose block-level architecture-oriented repairing for DNN repair. The intuition of block structure design in modern DNNs provides a suitable granularity of DNN repair at the block-level \cite{he2016resnet}. In addition, we also show that jointly optimizing architecture and weights further brings the advantage of DNN repair over repairing DNN by only updating weights, which is demonstrated by our comparative evaluation in the experimental section.

    \item In terms of the \textit{novelty and potential impacts}, existing DNN repair methods \cite{ma2018mode,zhang2019apricot,sohn2019arachne,gao2020sensei,eniser2019deepfault,ren2020fewshot} mostly focus on only repairing DNN via updating its weights while ignoring inherent DNN architecture design (\eg, block structure and relationships between different layers), which could also impact the DNN behavior, whereas only repairing the weights could not address such an issue. Therefore, compared with existing work, this paper initiates a new and wide direction for DNN repair by taking relationships of DNN architecture design as well as layers and weights into consideration.

    \item Technically, we originally propose the adversarial-aware spectrum analysis-based block localization and architecture-oriented search-based repairing method, both of which are novel for DNN repair. The first one enables us to localize a vulnerable block accurately even with only a few examples. The latter formulates the repairing problem as the joint optimization of both the architecture and weights at the block level.
    
    \item We implement our repairing techniques in the tool \method{} and perform extensive evaluation against 6 state-of-the-art DNN repair techniques under 4 DNNs with different architectures on two different datasets. The results demonstrate the advantage of \method{} in achieving SOTA repairing performance in terms of both accuracy and robustness. 
\end{itemize}
To the best of our knowledge, this is the very first attempt to consider the DNN repairing problem at the block-level that repairs both network weights and architecture jointly. The results of this paper demonstrate the limitation of repairing DNN by only updating the weights, and show that other important DNN development elements such as architecture that encodes more advanced relationships of neurons and layers should also be taken into consideration during the design of DNN repair techniques.

\section{DNN Repairing and Motivation}\label{sec:motive}

In this section, we review existing repairing methods in DNN and motivate our method. In \secref{subsec:motive-solution}, we thoroughly analyze previous DNN repair techniques from the viewpoint of different repairing targets, \eg, the parameters (\ie, weights) of the whole network, layers, or neurons. 
To this end, we formulate the core mechanism and compare their strengths and weaknesses, which inspires and motivates us to develop the block-level repairing method. 
To validate our motivation, we perform a preliminary study in \secref{subsec:motive-empirical}.

\subsection{DNN repair Solutions}\label{subsec:motive-solution}

In the standard training process, given a training dataset, we can train a DNN denoted as $\phi_{(\mathcal{W,}\mathcal{A})}$ where $\mathcal{A}$ represents the network architecture related parameters determining what operations (\eg, convolution layer, pooling layer, \etc) are used in the architecture, and $\mathcal{W}$ is the respective weights (\ie, parameters of different operations). Generally, the architecture $\mathcal{A}$ is pre-defined and fixed during the training and testing processes. The variable $\mathcal{W}$ consists of weights for different layers. 

Although existing DNNs (\eg, ResNet \cite{he2016resnet}) have achieved significantly high accuracy on popular datasets, incorrect behaviors are always found in these models when we deploy them in the real world or test them on challenging datasets. 
There are a series of works that study how to repair these DNNs to be generalizable to misclassified examples, challenging corruptions, or bias errors \cite{sohn2019arachne,ren2020fewshot,yu2021deeprepair,tian2020repairing}. 
In general, we can formulate the existing repairing methods as 
%
\begin{align}
\mathcal{W}^*
&=
\text{Locator}(\phi_{(\mathcal{W,}\mathcal{A})},\mathcal{D}^{\text{repair}})
\label{eq:unified_repair-1}
\\
\hat{\mathcal{W}}^{*}
&=
\argmin_{\mathcal{W}^*} \text{J}(\phi_{(\mathcal{W}^*,\mathcal{A})}, \mathcal{D}^{\text{repair}})
\label{eq:unified_repair-2}
%
\end{align}
%
where $\mathcal{W}^*$ is a subset of $\mathcal{W}$ and $\hat{\mathcal{W}}^*$ is the fixed counterpart of $\mathcal{W}^*$.
The dataset $\mathcal{D}^\text{repair}$ contains the examples for repairing guidance. Different works may set different $\mathcal{D}^\text{repair}$ according to the repairing scenarios.
For example, Yu \etal~\cite{yu2021deeprepair} sets $\mathcal{D}^\text{repair}$ as the combination of the augmented training dataset. 
We will show that our method can address different repairing scenarios.
Intuitively, \reqref{eq:unified_repair-1} is to find the weights we need to fix in the DNN, and \reqref{eq:unified_repair-2} with a task-related objective function $\text{J}(\cdot)$ is to fix the selected weights ${\mathcal{W}}^*$ and produce a new one $\hat{\mathcal{W}}^*$.

The above formulation can represent a series of existing repairing methods.
For example, when we try to fix all weights of a DNN (\ie, $\mathcal{W}^*=\mathcal{W}$) and set the objective function $\text{J}(\cdot)$ as the task-related loss function (\eg, cross-entropy function for image classification) with different data augmentation techniques on collected failure cases as $\mathcal{D}^\text{repair}$ to retrain the weights, we actually get the methods proposed by \cite{ren2020fewshot} and \cite{yu2021deeprepair}.
In addition, when we employ the gradient loss of weights and forward impact to localize the targeted weights and use a fitness function to fix localized weights, the formulation becomes the method \cite{sohn2019arachne}.

Nevertheless, with the general formulation in \reqref{eq:unified_repair-1} and \reqref{eq:unified_repair-2}, we can see that existing methods have the following limitations: 

\begin{itemize} 
    \item Existing works only fix the targeted DNN either at the network-level (\ie, fixing all weights of the DNN) or at the neuron-level (\ie, only fixing partial weights of the DNN), and ignore the effects of the architecture $\mathcal{A}$. 
    \item Only repairing some specific weights in a layer could easily neglect the relationship between different layers while repairing the whole network's weights leads to high cost. 
\end{itemize}

Note that, the state-of-the-art DNNs (\eg, ResNet \cite{he2016resnet}) are made up of several blocks where each block is built with stacked convolutional and activation layers. Such block-like architecture is mainly inspired by the philosophy of VGG nets \cite{Simonyan2015ICLR} and its effectiveness has been demonstrated in wide applications.  
Therefore in this work, we focus on DNN repairing at the block-level. In particular, we consider both the architecture and weights repairing of a specific block.

\subsection{Empirical Study and Motivation} \label{subsec:motive-empirical}

First, we perform a preliminary experiment to discuss the effectiveness of the repairing methods at different levels. In this experiment, we choose 3 variants of ResNet \cite{he2016resnet} (specifically, ResNet-18, ResNet-50, and ResNet-101) as the targeted DNNs $\phi$, and we select CIFAR-10 dataset as the experimental environment. We repair the DNN at four levels, \ie, Neuron-level (\ie, only fixing weights of one neuron ), Layer-level (\ie, only fixing the weights of one layer), Block-level (\ie, fixing the weights of a block) and the Network-level (\ie, fixing all weights of the DNN). 
Inspired by recent work \cite{sohn2019arachne}, we choose the neuron (or layer/block) with the greatest gradient (mean gradient for layer and block) as our target to fix. Note that as the previous work have shown that repairing DNN with only a few failure cases is meaningful and important~\cite{ren2020fewshot,yu2021deeprepair}, we only randomly select 100 failure cases from the testing dataset to calculate the gradients and choose such neuron (or layer/block). 
Then, we adjust the weights of the chosen neuron/layer/block by gradient descent \wrt the loss function (\eg, cross-entropy loss for image classification).
To compare their effectiveness, we apply all methods on the same training dataset of CIFAR-10 and Tiny-ImageNet, then measure the accuracy on the respective testing dataset. We also record the execution time of the total repairing phase (100 epochs) as indicator of time cost. We show the repairing result in \tableref{tab:motivate-empirical}.

According to \tableref{tab:motivate-empirical}, the network-level repairing achieves the highest accuracy on ResNet-18 and ResNet-101 when repairing CIFAR-10 dataset, and all 3 variants of ResNet when repairing Tiny-ImageNet dataset, but also leads to the highest time cost under every configuration. 
Among 3 other levels of repairing methods, the block-level repairing achieves the highest accuracy improvement without having drastic increment on time cost (\ie, the run-time increment comparing with neuron-level and layer-level is less than 500 seconds on 100 epochs across all 3 ResNets) when repairing on both CIFAR-10 and Tiny-ImageNet. 

\begin{table*}[t]
    \centering
    \small
    \caption{Accuracy (\%) and execution time (s/100 epochs) of applying repairing method at different levels on 3 different DNNs trained and tested on CIFAR-10 and Tiny-ImageNet datasets. }
    {\resizebox{\linewidth}{!}{
    \begin{tabular}{c|l|rr|rr|rr}
    \toprule
        \multicolumn{2}{c|}{\multirow{2}{*}{\bf Scale}} & \multicolumn{2}{c|}{\bf ResNet-18} & \multicolumn{2}{c|}{\bf ResNet-50} & \multicolumn{2}{c}{\bf ResNet-101}\\
        \multicolumn{2}{c|}{} & \bf Accuracy (\%) & \bf Execution Time & \bf Accuracy (\%) & \bf Execution Time & \bf Accuracy (\%) & \bf Execution Time \\
    \midrule
        \multirow{5}{*}{\rotatebox{90}{CIFAR-10}}
        & \bf Original        & 85.00    & - & 85.17    & - & 85.31    & - \\
        & \bf Neuron-level    & 85.18 & 650.49 & 85.23 & 4054.29 & 85.39 & 6853.47 \\
        & \bf Layer-level     & 85.16 & 590.47 & 85.24 & 4159.93 & 85.41 & 4956.81 \\
        & \bf Block-level     & 85.19 & 760.94 & 85.24 & 3976.39 & 85.47 & 7118.03 \\
        & \bf Network-level   & 85.73 & 1456.92 & 84.80 & 5735.61 & 87.43 & 9889.35 \\
    \midrule
        \multirow{5}{*}{\rotatebox{90}{Tiny-ImageNet}}
        & \bf Original        & 45.15    & - & 46.26    & - & 46.14    & - \\
        & \bf Neuron-level    & 45.23 & 1847.59 & 46.17 & 13074.85 & 46.14 & 20395.79 \\
        & \bf Layer-level     & 45.23 & 1854.37 & 46.24 & 12796.91 & 46.15 & 18497.53 \\
        & \bf Block-level     & 45.30 & 2011.84 & 46.27 & 13452.17 & 46.22 & 24774.15 \\
        & \bf Network-level   & 45.52 & 2574.81 & 46.41 & 17495.88 & 46.55 & 32908.43 \\
    \bottomrule
    \end{tabular}
    }
    }
    \label{tab:motivate-empirical}
\end{table*}

Overall, the network-level repairing is significantly effective on the accuracy improvement but leads to a high time cost. 
Nevertheless, the block-level repairing achieves impressive accuracy enhancement with much less execution time comparing to network-level method (\eg, about $2\times$ less on ResNet-18), making it a good trade-off between effectiveness and efficiency. 
This fact inspires and motivates us to further investigate the block-level repairing method.

\section{Block-level Architecture and Weights Repairing}\label{sec:blrepair}

In this section, we first provide an overview of our method in the \secref{subsec:blr-overview} by presenting our intuitive idea and the main pipeline containing two key modules, \ie, \textit{Vulnerable Block Localization} and \textit{Architecture-oriented Search-based Repairing}. After that, we detail the first module in \secref{subsec:blr-locating} and the second module in \secref{subsec:blr-searching}, respectively. The first module is to locate the vulnerable block in a deployed DNN, while the second module is to repair the architecture and weights of the localized block by formulating it as an architecture searching problem. 

\subsection{Overview}\label{subsec:blr-overview}

Given a deployed DNN $\phi_{(\mathcal{W},\mathcal{A})}$, the weights and architecture usually consist of several blocks, each of which is built by stacking basic operations, \eg, convolutional layer. 
Then, we represent the weights and architecture with $B$ blocks, \ie, $\mathcal{W} = \{\mathcal{W}_{\text{b}}^i\}_{i=1}^{B}$ and $\mathcal{A} = \{\mathcal{A}_{\text{b}}^i\}_{i=1}^{B}$, where the weights or architecture of each block are made up by one or multiple layers. 
For example, when we consider the ResNet-18 \cite{he2016resnet}, we can say that it has six blocks (See \tableref{tab:resnet-block}). The first block contains only one convolution layer with the kernel size of $7\times 7 \times 64$ and the stride of 2. The second to the fifth blocks have two convolutional layers and the last block contains a fully connected layer and a softmax layer. 
Then, we can reformulate \reqref{eq:unified_repair-1} and \reqref{eq:unified_repair-2} for the proposed block-level repairing by
%
\begin{align}
%
(\mathcal{W}_\text{b}^*,\mathcal{A}_\text{b}^*) &= \text{Locator}(\phi_{(\{\mathcal{W}_{\text{b}}^i\}_{i=1}^{B},\{\mathcal{A}_{\text{b}}^i\}_{i=1}^{B})},\mathcal{D}^{\text{repair}}) \label{eq:block_repair-1} \\
(\hat{\mathcal{W}}^{*}_\text{b}, \hat{\mathcal{A}}^{*}_\text{b}) &= \argmin_{(\mathcal{W}_\text{b}^*,\mathcal{A}_\text{b}^*)} \text{J}(\phi_{(\mathcal{W}_\text{b}^*,\mathcal{A}_\text{b}^*)}, \mathcal{D}^{\text{repair}}) \label{eq:block_repair-2}
%
\end{align}
%
where \reqref{eq:block_repair-1} is to locate the block (\ie, $(\mathcal{W}_\text{b}^*,\mathcal{A}_\text{b}^*)$) that should be fixed through the proposed adversarial-aware block localization, and \reqref{eq:block_repair-2} is to repair the localized block by formulating it as a network architecture searching problem. Clearly, compared with the general repairing method (\ie, \reqref{eq:unified_repair-1} and \reqref{eq:unified_repair-2}), the proposed method focuses on fixing the weights and architecture at the block level. 
We detail the \textit{vulnerable block localization} in \secref{subsec:blr-locating} and \textit{architecture search-based repairing} in \secref{subsec:blr-searching}.

\begin{table}[t]
    \setlength{\tabcolsep}{3pt}
    \centering
    \caption{ResNet architectures and their respective blocks. More details could be found in \cite{he2016resnet}.}
    \begin{tabular}{ll|ccc}
    \toprule
        Block & Layer & 18-layer & 50-layer &  101-layer \\
    \midrule
        Blk1  & conv1 & \multicolumn{3}{c}{$7\times 7$, 64, stride 2}\\
        \hline
        \multirow{2}{*}{Blk2}  & \multirow{2}{*}{conv2\_x} & \multicolumn{3}{c}{$3\times 3$ max pool, stride 2}\\
        \cline{3-5}
        & & $\begin{bmatrix} 3\times 3,64\\ 3\times 3,64 \end{bmatrix}\times2$ & $\begin{bmatrix} 1\times 1,64\\ 3\times 3,64\\ 1\times 1,256\\ \end{bmatrix}\times3$ & $\begin{bmatrix} 1\times 1,64\\ 3\times 3,64\\ 1\times 1,256\\ \end{bmatrix}\times3$ \\
        \hline
        Blk3  & conv3\_x & $\begin{bmatrix} 3\times 3,128\\ 3\times 3,128 \end{bmatrix}\times2$ & $\begin{bmatrix} 1\times 1,128\\ 3\times 3,128\\ 1\times 1,512\\ \end{bmatrix}\times4$ & $\begin{bmatrix} 1\times 1,128\\ 3\times 3,128\\ 1\times 1,512\\ \end{bmatrix}\times4$ \\
        \hline
        Blk4  & conv4\_x & $\begin{bmatrix} 3\times 3,256\\ 3\times 3,256 \end{bmatrix}\times2$ & $\begin{bmatrix} 1\times 1,256\\ 3\times 3,256\\ 1\times 1,1024\\ \end{bmatrix}\times6$ & $\begin{bmatrix} 1\times 1,256\\ 3\times 3,256\\ 1\times 1,1024\\ \end{bmatrix}\times23$ \\
        \hline
        Blk5  & conv5\_x & $\begin{bmatrix} 3\times 3,512\\ 3\times 3,512 \end{bmatrix}\times2$ & $\begin{bmatrix} 1\times 1,512\\ 3\times 3,512\\ 1\times 1,2048\\ \end{bmatrix}\times3$ & $\begin{bmatrix} 1\times 1,512\\ 3\times 3,512\\ 1\times 1,2048\\ \end{bmatrix}\times3$ \\
        \hline
        Blk6  &  \multicolumn{4}{c}{average pool, 1,000-d fully-connection, softmax}\\
    \bottomrule
    \end{tabular}
    \label{tab:resnet-block}
\end{table}

\begin{figure}[t]
    \centering
    \includegraphics[width=0.8\linewidth]{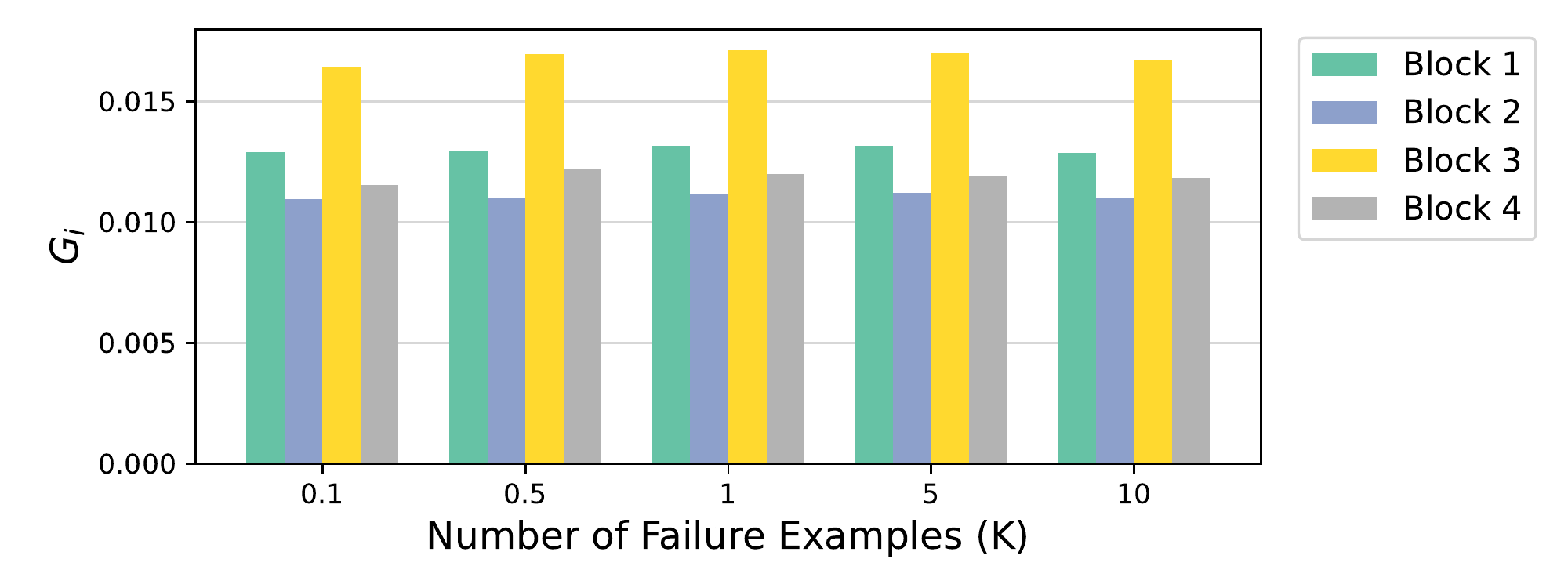}
    \caption{
       Average gradients of different blocks in ResNet-18 for different $\mathcal{D}^{\text{repair}}_{\text{fail}}$ sizes.
    }
    \label{fig:gr_bb}
\end{figure}

There are two main solutions for vulnerable neurons localization \cite{sohn2019arachne,eniser2019deepfault}. The first one employs the neuron spectrum analysis during the forward process of DNN on a testing dataset. It calculates the spectrum of all neurons (\eg, activated/non-activated times of neurons for correctly classified examples and activated/non-activated times of neurons for misclassified examples). 
These attributes are used to measure the suspiciousness of all neurons. The general principle is that a neuron is more suspicious when the neuron is more often activated under the misclassified examples than that under the correctly classified examples \cite{eniser2019deepfault}.
This solution is able to localize the vulnerable neurons accurately but requires a large testing dataset, which is not suitable for the scenario where a few examples are available for repairing. 
The second solution is to actively localize the vulnerable neurons by performing backpropagation on the misclassified examples and calculating the gradients of neurons \wrt the loss function. The neurons with large gradients are responsible for the misclassification \cite{sohn2019arachne}.
This solution is able to localize the vulnerable neuron with fewer examples but ignores the effects of correctly classified examples. As shown in \figref{fig:gr_bb}, with different failure examples, the gradients of different convolutional blocks in ResNet18 may have similar values, which demonstrates that the gradient-based localization is not sensitive to the variance of the number of failure examples.

Overall, existing methods only focus on localizing vulnerable neurons while ignoring the blocks in DNNs. In addition, they have their respective defects.
In this work, we propose a novel localization method that aims to find the most vulnerable block in the DNN, which can lead to the buggy behavior of a deployed DNN.
To take the respective advantages of existing works and avoid their defects, we propose adversarial-aware spectrum analysis to localize the vulnerable block.

\subsection{Adversarial-aware Specturm Analysis for Vulnerable Block Localization}\label{subsec:blr-locating}

\subsubsection{Neuron spectrum  analysis}
Given a dataset $\mathcal{D}^\text{repair}$ for repairing and the targeted DNN $\phi_{(\mathcal{W},\mathcal{A})}$, we calculate the spectrum attributes of the $j$th neuron in $\mathcal{W}$ by counting the times of activation and non-activation for the neuron under the correctly classified examples and denote them as $N^j_{\text{ac}}$ and $N^j_{\text{nc}}$, respectively. Similarly, we can count the times of activation and non-activation for the same neuron under the misclassified examples and name them as $N^j_{\text{am}}$ and $N^j_{\text{nm}}$, respectively.
Then, we calculate a suspiciousness score for each neuron via the Tarantula measure \cite{jones2005tarantula}, 
%
%
\begin{align}\label{eq:tarantula}
s_j = \frac{N^j_{\text{am}}/(N^j_{\text{am}}+N^j_{\text{nm}})}{N^j_{\text{am}}/(N^j_{\text{am}}+N^j_{\text{nm}})+N^j_{\text{ac}}/(N^j_{\text{ac}}+N^j_{\text{nc}})} 
\end{align}
%
where $s_j$ determines the suspiciousness of the $j$th neuron and the higher $s_j$ means the $j$th neuron is more vulnerable.

\subsubsection{Adversarial-aware block spectrum analysis}
With the above neuron spectrum analysis, we can obtain the suspiciousness scores for all neurons and the suspiciousness set $\mathcal{S}=\{s_j\}$. 
Nevertheless, these suspiciousness scores depend on the statistical analysis and are not related to the objective directly, which leads to less effective localization. 
To alleviate the issue, we propose to refine the suspiciousness scores with adversarial information under the guidance of the loss function (\eg, cross-entropy function for classification).

Specifically, we select the failure examples in $\mathcal{D}^\text{repair}$ and construct a subset denoted as $\mathcal{D}^\text{repair}_\text{fail}$. 
For each example in $\mathcal{D}^\text{repair}_\text{fail}$, we can calculate the gradient of all neurons \wrt the loss function. Then, we average the gradients of a neuron on all examples and get a set $\mathcal{G} = \{g_j\}$ where $g_j$ is the averaging gradient of the $j$th neuron on all examples in $\mathcal{D}^\text{repair}_\text{fail}$. 
Intuitively, the larger gradient means that the corresponding neuron may significantly contribute to misclassification and should be tuned to minimize the loss. 
For the $i$th block, we denote its gradient as the average of the gradients of all neurons in that block, \ie, $G_i = \frac{1}{|\mathcal{W}_\text{b}^i|}\sum_{\mathbf{w}_j\in\mathcal{W}_\text{b}^i}g_j$.
We also calculate the averaging gradient across all neurons, \ie, $\overline{G}=\frac{1}{B}\sum_{i=1}^{B}G_i$.
Then, we use these gradients to reweight the suspiciousness scores of all neurons.
%
\begin{align}\label{eq:reweight}
\hat{s}_j = \frac{|g_j-\overline{G}|}{\max(\{|g_j-\overline{G}|\})} s_j.
\end{align}
%
The principle behind this strategy is that the suspiciousness score of the $j$th neuron decreases when its relative gradient is small.
As a result, we can update the suspiciousness set $\mathcal{S}$ and get $\hat{\mathcal{S}}=\{\hat{s}_j\}$.

A block in the DNN consists of a series of neurons and we collect the updated suspiciousness scores of the neurons in the $i$th block to the set $\hat{\mathcal{S}}_i\in\hat{\mathcal{S}}$.
There are $B$ suspiciousness sets and $\hat{\mathcal{S}} = \{\hat{\mathcal{S}}_i\}_{i=1}^B$.
After that, we use a threshold (\ie, $\epsilon$) to select the vulnerable neurons, that is, the neuron with $\hat{s}_j>\epsilon$ is identified as the vulnerable neuron. 
Then, we can count the number of vulnerable neurons in each $\hat{\mathcal{S}}_i$ and the block with the most vulnerable neurons is identified as the targeted block we would repair.

\begin{algorithm}[t]
\caption{Vulnerable block localization}
\label{alg:suspiciousness_ranking}
\KwIn{ A DNN $\phi_{(\mathcal{W},\mathcal{A})}$ and datasets $\mathcal{D}^\text{repair}$ and $\mathcal{D}^\text{repair}_\text{fail}$}
\KwOut{$\mathcal{W}^*_\text{b}$,$\mathcal{A}^*_\text{b}$}
Calculate suspiciousness scores $\mathcal{S}$ of all neurons via \reqref{eq:tarantula}\;
Calculate the gradients of all neurons on $\mathcal{D}^\text{repair}_\text{fail}$ and get $\mathcal{G}$\;
Update the suspiciousness scores $\mathcal{S}$ and get $\hat{\mathcal{S}}$\;
Identify the vulnerable neurons via a threshold $\epsilon$\;
Localize the vulnerable block with maximum number of vulnerable neurons\;
\end{algorithm}

\begin{figure}[t]
    \centering
    \includegraphics[width=0.8\linewidth]{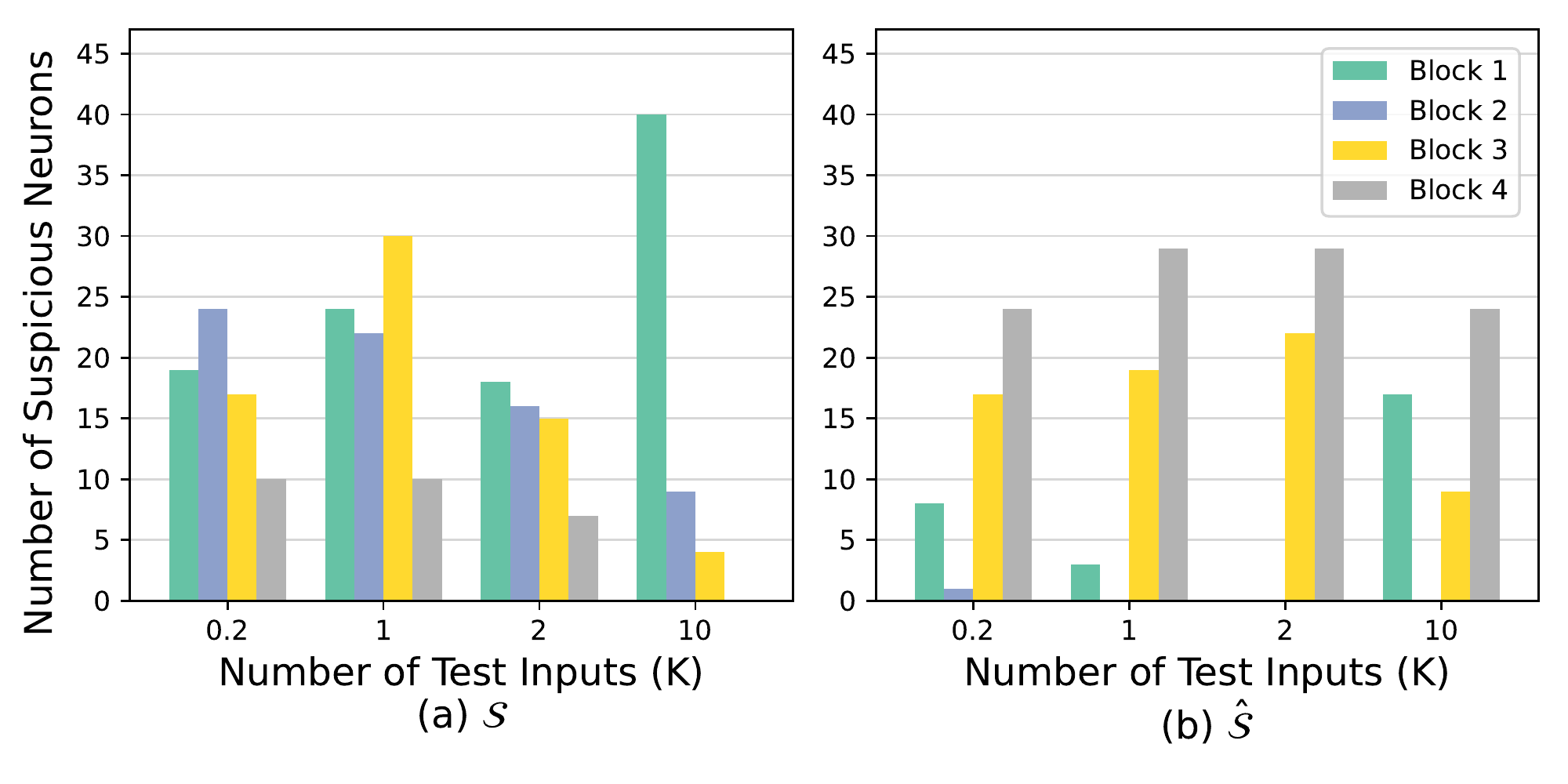}
    \caption{
        Collected suspicious neurons in blocks of ResNet-18 when setting threshold $\epsilon$ equals to the value that select top-$50$ neurons from suspicious ranking, with $\mathcal{S}$(left) and $\hat{\mathcal{S}}$(right), respectively. 
    }
    \label{fig:sr_comparsion}
\end{figure}

We summarize the whole process of the block localization in Algorithm~\ref{alg:suspiciousness_ranking}.
To validate its advantages, we conduct an experiment to compare the effectiveness and stability of the blocks positioned from ${\mathcal{S}}$ and $\hat{\mathcal{S}}$, respectively. 
To compare the stability of the method, we changed the size of the dataset $\mathcal{D}^{\text{repair}}_\text{fail}$. 
We observe that as the size of the dataset changes, the suspicious neurons on each block obtained by ${\mathcal{S}}$ vary significantly while those obtained by $\hat{\mathcal{S}}$ are much more stable and lead to unanimous conclusions. 
As shown in \figref{fig:sr_comparsion}, according to the experiments on ResNet-18, by the number of suspicious neurons contained in the block, ${\mathcal{S}}$ and $\hat{\mathcal{S}}$ estimated that `block 1' and `block 4' are the most vulnerable, respectively. We observed similar results when the threshold $\epsilon$ are set to other values (\eg, $\epsilon_{10}$, $\epsilon_{20}$, $\epsilon_{30}$, $\epsilon_{40}$, $\epsilon_{100}$).
We also conduct detailed quantitative analysis and discussion in \secref{subsec:exp-rq3}, presenting that repairing the most vulnerable block, \ie, `block 4', achieves much higher improvement. 

\subsection{Architecture-oriented Search-based Repairing}\label{subsec:blr-searching}

After localizing the targeted block, how to break the old architecture's bottleneck and fix it to become competent in the tasks is another challenge.
To this end, we formulate the very first block-level architecture and weights repairing as the network architecture search task.
Given a deployed DNN with pre-trained weights and fixed architecture (\ie, $\phi_{(\mathcal{W},\mathcal{A})}$), we first relax the targeted block (\ie, $\phi_{(\mathcal{W}^*_\text{b},\mathcal{A}^*_\text{b})}$) to a directed acyclic graph like the cell structure in the differentiable architecture search (DARTS) \cite{liu2018darts}, which is composed of an ordered sequence of nodes that are connected by edges. Intuitively, the node corresponds to the deep feature while the edge denotes the operation layer like convolutional layer. 
Our goal is to optimize the edges, \ie, to determine which two nodes should be connected and which operation should be selected for that connection.
To this end, the key issues are to define the architecture search space and optimization strategy.

\subsubsection{Architecture search space for the targeted block} 
\begin{figure*}[t]
    \centering
    \includegraphics[width=\linewidth]{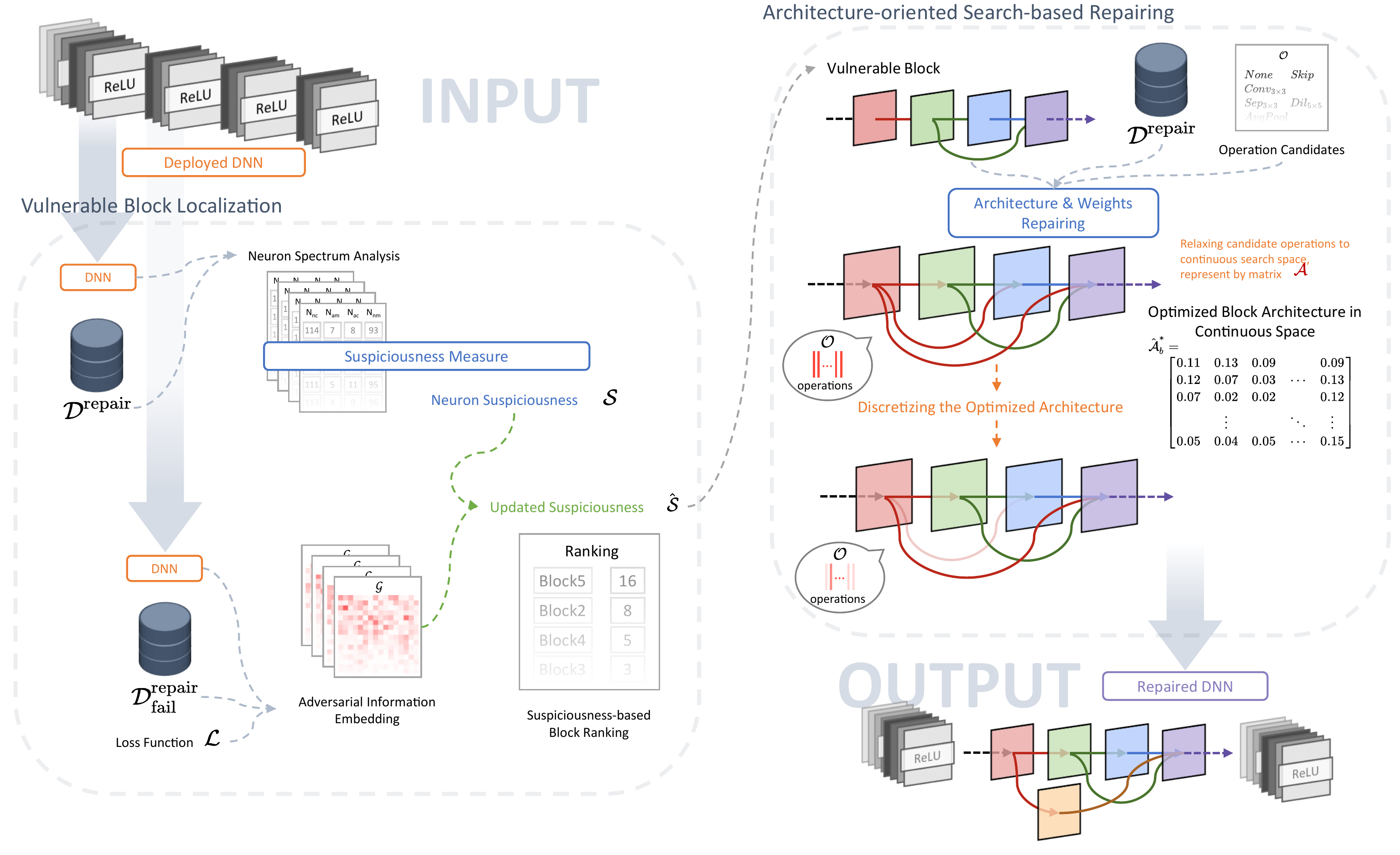}\\
    \caption{The overall workflow of \method{}. Given a deployed DNN model, we first apply the Vulnerable Block Localization to identify the most vulnerable block. Then, we continue to formulate the block repairing as a DNN architecture search problem, and the block's architecture and parameters are optimized jointly through Architecture-oriented Search-based Repairing.}
    \label{fig:workflow}
\end{figure*}

To better illustrate the process of architecture search, we take the ResNet as an example. Given a block in ResNet containing $K$ operation layers, we reformulate it as a directed acyclic graph that has $K+1$ nodes $\{\mathbf{X}^k\}_{k=1}^{K}$ and allow each node to accept the outputs from all previous nodes instead of following the sequential order. 
As shown in \figref{fig:workflow}, we present an example of the graph representation of the targeted block via nodes and edges. 
Specifically, we denote the edge for connecting the $i$th and $j$th nodes as $\text{e}_{(i,j)}$ and the node $\mathbf{X}^j$ can be calculated by
%
\begin{align} \label{eq:cal_node}
\mathbf{X}^j=\sum_{i=[1,j-1]}\text{e}_{(i,j)}(\mathbf{X}^{i}),
\end{align}
%
where $\text{e}_{(i,j)}(\mathbf{X}^{i})$ is an edge taking the node $\mathbf{X}^{i}$ as the input.
Then, we define an operation set $\mathcal{O}$ containing six candidate operations as presented in \tableref{tab:op}, each of which can be set as the edge. 
For example, when we select `None' for $\text{e}_{(i,j)}$, the two nodes $\mathbf{X}^{i}$ and $\mathbf{X}^{j}$ should not be connected.

Note that, the raw sequentially ordered block of ResNet is a special case in the defined search space and we can naturally inherent the raw weights and architecture setup as the initialization for the following optimization.

\subsubsection{Architecture and weights optimization}
The optimization goal is to select a suitable operation for each edge from the operation set.
To this end, we relax the selection as a continuous process by regarding the edge connecting the node $i$ and $j$ as a weighted combination of the outputs of all candidate operations
%
\begin{align} \label{eq:relax_node}
\text{e}_{(i,j)}(\mathbf{X}^{i})=\sum_{\text{o}\in\mathcal{O}}\frac{\exp{(\alpha^\text{o}_{(i,j)})}}{\sum_{\text{o}'\in\mathcal{O}} \exp{(\alpha^{\text{o}'}_{(i,j)})}}\text{o}(\mathbf{X}^{i})
\end{align}
%
where the parameter $\alpha^\text{o}_{(i,j)}$ determines the combination weight of using the operation $\text{o}$ for connecting the $i$th and $j$th nodes. 
As a result, we can define the architecture parameters for the edge $\text{e}_{(i,j)}$ as a vector $\mathbf{a}_{(i,j)}=[\alpha_{(i,j)}^\text{o}|\text{o}\in\mathcal{O}]$ assigning each operation in the $\mathcal{O}$ a combination weight.
Moreover, for the whole block, we denote its architecture as $\mathcal{A}_{\text{b}}^*=\{\mathbf{a}_{(i,j)}\}$ and respective parameters for all candidate operations as $\mathcal{W}_{\text{b}}^*=\{\mathbf{w}_{(i,j)}\}$.
Then, we can specify the repairing process in \reqref{eq:block_repair-2} by optimizing the weights (\ie, $\mathcal{W}_\text{b}^*$) and architecture parameters  (\ie, $\mathcal{A}_\text{b}^*$) on the training dataset and validation dataset, alternatively, that is, we have
%
\begin{align}
\hat{\mathcal{W}}^{*}_\text{b} &= \argmin_{\mathcal{W}_\text{b}^*} \text{J}(\phi_{(\mathcal{W}_\text{b}^*,\mathcal{A}_\text{b}^*)}, \mathcal{D}^{\text{repair}}_{\text{train}}) \label{eq:nas_repair-1}, \\
\hat{\mathcal{A}}^{*}_\text{b} &= \argmin_{\mathcal{A}_\text{b}^*} \text{J}(\phi_{(\hat{\mathcal{W}}_\text{b}^*,\mathcal{A}_\text{b}^*)}, \mathcal{D}^{\text{repair}}_{\text{val}}) \label{eq:nas_repair-2}
\end{align}
%
where $\text{J}(\cdot)$ is specified as the cross-entropy loss function for the image classification task.
During the training process, we initialize the block architecture $\mathcal{A}^{*}_\text{b}$ as the raw block architecture of the targeted DNN, and update the architecture and weights, alternatively.
We will detail the repairing process in \secref{subsec:blr-workflow}.
After getting the optimized architecture (\ie, $\hat{\mathcal{A}}_\text{b}^*$) in the continuous search space, we set the operation with maximum combination weight as the edge, \ie, $\text{e}_{(i,j)} =\argmax_{\text{o}\in\mathcal{O}}\alpha_{(i,j)}^\text{o}$. Then, we retrain the weights $\hat{\mathcal{W}}^*_\text{b}$ with fixed block architecture.

\begin{table}[t]
    \centering
    \caption{All operators in the operation set $\mathcal{O}$.}
    {
    \begin{tabular}{l|l}
    \toprule
        Operators & Operations \\
    \midrule
        None        & Add a Zero CNN layer whose weights are all zero. \\
        Skip        & Add an Identity CNN layer whose weights are all one. \\
        AvgPool     & Add an Average Pooling layer and an Identity CNN layer. \\
        MaxPool     & Add a Max Pooling layer and an Identity CNN layer. \\
        SepConv     & Add separated CNN layers. \\
        DilConv     & Add a CNN layer with dilation kernel and an Identity CNN layer. \\
    \bottomrule
    \end{tabular}
    }
    \label{tab:op}
\end{table}

\subsection{Our Repairing Algorithm}
\label{subsec:blr-workflow}

\figref{fig:workflow} displays the whole workflow of \method{}. 
Given a deployed DNN, we first employ the proposed vulnerable block localization to determine the block we aim to fix. 
Specifically, we use the $\mathcal{D}^\text{repair}$ dataset and the neuron spectrum analysis to get the suspiciousness scores of all neurons, \ie, $\mathcal{S}=\{s_j\}$. 
Meanwhile, we use the failure examples in $\mathcal{D}^\text{repair}$ (\ie, $\mathcal{D}^\text{repair}_\text{fail}$) to get the gradients of all neurons \wrt the loss function (\ie, $\mathcal{G}=\{g_j\}$). 
Then, we use Eq.~\eqref{eq:reweight} and $\mathcal{G}=\{g_j\}$ to reweight $\mathcal{S}=\{s_j\}$, thus get $\hat{\mathcal{S}}=\{\hat{s}_j\}$.
After that, we can calculate the number of vulnerable neurons through a threshold $\epsilon$, that is ,when the suspiciousness score of a neuron is larger than $\epsilon$, the neuron is identified as a vulnerable case. 
Finally, the block with the largest number of vulnerable cases is selected as the targeted block we want to repair.

During the architecture search-based repairing, we reformulate the targeted block as a directed acyclic graph where the deep features are nodes and operations are edges. Then, we relax each edge as a combination of six operations (\ie, \reqref{eq:relax_node}) where the combination weights correspond to the architecture parameters $\mathcal{A}_\text{b}^*=\{\mathbf{a}_{(i,j)}\}$. We use the dataset $\mathcal{D}^\text{repair}$ to conduct the architecture and weights optimization via \reqref{eq:nas_repair-1} and \reqref{eq:nas_repair-2} where the original architecture and weights are inherited and serve as the optimization initialization.
Hence given the optimized block architecture in the continuous space (\ie, $\hat{\mathcal{A}}^*_\text{b}$), we discretize it to the final architecture by preserving the operation with the maximum combination weight and removing other operations.
Finally, we use the $\mathcal{D}^\text{repair}$ to fine-tune the weights by fixing the optimized architecture for the repaired DNN.

\section{Experimental Design and Settings}\label{sec:exp}
    
In this section, we conduct extensive experiments to validate the proposed methods and compare with the state-of-the-art DNN repair techniques, to investigate the following research questions:

\begin{itemize} 
    \item \textbf{RQ1.} Does \method{} outperform the state-of-the-art (SOTA) DNN repair techniques with better repairing effects?
    
    \item \textbf{RQ2.} Could \method{} repair DNNs on certain failure patterns without sacrificing robustness on clean data and other failure patterns?
    
    \item \textbf{RQ3.} Is our proposed localization method effective in identifying vulnerable neuron blocks?
    
    \item \textbf{RQ4.} How do different components of our proposed method impact the overall repairing performance?
\end{itemize}

\textbf{RQ1} intends to evaluate the overall repairing capability of \method{} and to compare it to SOTA DNN repair techniques as baselines. \textbf{RQ2} aims at exploring the potential of our method in repairing DNN on corrupted data, which are common robustness issues during DNN practical usage in the operational environments. \textbf{RQ3} intends to examine whether the proposed localization method can precisely locate vulnerable blocks. \textbf{RQ4} is to explore the contribution that each of \method{}'s key components makes on the overall performance of DNN repair.

\subsection{Experimental Setups} \label{subsec:exp-setup}

To answer the research questions above, we design our evaluation from multiple perspectives listed in the following.

{\bf Subject Datasets and Repairing Scenarios.}
Given a deployed DNN trained on a training dataset $\mathcal{D}^\text{t}$, we can evaluate it on a testing dataset $\mathcal{D}^\text{v}$. 
In the real world, there are a lot of scenes that cannot be covered by $\mathcal{D}^\text{v}$ and the DNN's performance may decrease significantly after the DNN is deployed in its operational environment.
For example, there are common corruptions (\ie, noise patterns) in the real world that can affect the DNN significantly~\cite{hendrycks2019robustness}: Gaussian noise (GN), shot noise (SN), impulse noise (IN), defocus blur (DB), Gaussian blur (GB), motion blur (MB), zoom blur (ZB), snow (SNW), frost (FRO), fog (FOG), brightness (BR), contrast (CTR), elastic transform (ET), pixelate (PIX), and JPEG compression (JPEG). 

According to above situations, we consider two repairing scenarios that commonly occur in practice: 
\begin{itemize} 
    \item {\bf Repairing the accuracy drift on testing dataset.} When we evaluate the DNN on the testing dataset $\mathcal{D}^\text{v}$, we can collect a few failure examples (\ie, 1,000 examples) denoted as $\mathcal{D}^\text{v}_{\text{fail}}$. Then, we set $\mathcal{D}^{\text{repair}}=\mathcal{D}^\text{v}_{\text{fail}}\cup\mathcal{D}^\text{t}$ and use the proposed or baseline repairing methods to enhance the deployed DNNs. We evaluate the accuracy on the testing dataset where $\mathcal{D}^\text{v}_{\text{fail}}$ is excluded (\ie, $\mathcal{D}^\text{v}\setminus\mathcal{D}^\text{v}_{\text{fail}}$). 
    Note that, the context of repairing DNN with only a few testing data is meaningful and important, which is adopted by recent works~\cite{ren2020fewshot,yu2021deeprepair}.
    In addition, there could be many practical scenarios in which collecting buggy example is very difficult or at very high cost, with only a few buggy examples collected entirely. Hence, we follow the common choice in recent works~\cite{ren2020fewshot,yu2021deeprepair} to select only 1,000 failure examples from testing data.
    \item {\bf Repairing the robustness on corrupted datasets.} When we evaluate the DNN on a corrupted testing dataset $\mathcal{D}^\text{c}$, we can also collect a few failure examples (\ie, 1,000 examples) denoted as $\mathcal{D}^\text{c}_{\text{fail}}$ and set $\mathcal{D}^{\text{repair}}=\mathcal{D}^\text{c}_{\text{fail}}\cup\mathcal{D}^\text{t}$. The repairing goal is to enhance the accuracy on $\mathcal{D}^\text{c}\setminus\mathcal{D}^\text{c}_{\text{fail}}$ and other corrupted datasets while maintaining the accuracy on the clean testing dataset (\ie, $\mathcal{D}^\text{v}\setminus\mathcal{D}^\text{v}_{\text{fail}}$). 
\end{itemize}

We choose CIFAR-10~\cite{krizhevsky2009cifar} and Tiny-ImageNet~\cite{le2015tinyimg} as the evaluation datasets. They are commonly used datasets in recent DNN repair studies, which enables us for comparative studies in a relatively fair way.
Each dataset contains their respective training dataset $\mathcal{D}^\text{t}$ and testing dataset $\mathcal{D}^\text{v}$.
CIFAR-10 contains a total of 60,000 images in 10 categories, in which 50,000 images are for $\mathcal{D}^\text{t}$ and the other 10,000 are for $\mathcal{D}^\text{v}$. 
Tiny-ImageNet has a training dataset $\mathcal{D}^\text{t}$ with the size of 100,000 images, and a testing dataset $\mathcal{D}^\text{v}$ with the size of 10,000 images.
Therefore, we have corrupted testing datasets $\{\mathcal{D}^\text{c}_i\}$ where $i=1,2,\dots, 15$ corresponding to the above fifteen corruptions~\cite{hendrycks2019robustness}.

{\bf DNN architectures.}
We select four different architectures of DNN, \ie, ResNet-18, ResNet-50, ResNet-101~\cite{he2016resnet}, and DenseNet-121~\cite{huang2017densenet}. Given that \method{} is a block-based repairing method, the block-like architecture, ResNet, turns out to be a perfect research subject. For a broad comparison, we also choose a non-block-like architecture, DenseNet-121, to examine the repairing capability of \method{}~\footnote{For DenseNet-121, we manually group two consecutive convolution blocks as one block when repairing.}. For each architecture, we first pre-train them with the original training dataset $\mathcal{D}^\text{t}$ (from CIFAR-10 or Tiny-ImageNet), the model with the highest accuracy in testing dataset $\mathcal{D}^\text{v}$ (from CIFAR-10 or Tiny-ImageNet) will be saved as pre-trained model $\phi_\theta$. As the original ResNet and DenseNet are not designed for CIFAR-10 and Tiny-ImageNet datasets, we use the unofficial architecture code offered by a popular GitHub project\footnote{Train CIFAR10 with PyTorch: https://github.com/kuangliu/pytorch-cifar}, which has more than 4.1K stars. 

{\bf Hyper-parameters.}
In terms of the training setup, we employ stochastic gradient descent (SGD) as the optimizer, setting batch size as 128, the initial learning rate as 0.1 and the weight decay as 0.0005. We use cross-entropy loss as the loss function. The maximum number of epochs is 500, and an early-stop function will terminate the training phase when the validation loss no longer decreases in 10 epochs. 

{\bf Baselines.}
To demonstrate the repairing capability of the proposed \method{}, we select 6 SOTA DNN repair methods from two different categories as baselines: neuron-level repairing methods and network-level repairing methods. The neuron-level repairing methods focus on repairing certain neurons' weight in order to repair the DNNs, representative methods from this category are MODE~\cite{ma2018mode}, Apricot~\cite{zhang2019apricot}, and Arachne~\cite{sohn2019arachne}. While network-level repairing methods mainly repair DNNs by using augmented datasets to fine-tune the whole network, where SENSEI~\cite{gao2020sensei}, Few-Shot~\cite{ren2020fewshot}, and DeepRepair~\cite{yu2021deeprepair} are the most popular ones. 
For a fair comparison, we employ the same settings on all six repairing methods and \method{}. In order to fully evaluate the effectiveness of proposed method, we apply all methods (six baselines and \method{}) to fix 4 different DNN architectures on large-scale datasets, including the clean version and 15 corrupted version from CIFAR-10 and Tiny-ImageNet, to assess the repairing capability.

{\bf Other configurations.}
We implement \method{} in Python 3.9 based on PyTorch framework. All the experiments were performed on a same server with a 12-core 3.60GHz Xeon CPU E5-1650, 128GB RAM and four NVIDIA GeForce RTX 3090 GPUs (24GB memory of each). The opreation system is Ubuntu 18.04. 

In summary, for each baseline method and \method{}, our evaluation consists of 64 configurations (4 DNN architectures $\times$ 16 versions of a dataset~\footnote{one clean dataset (repairing the accuracy drift on testing dataset) and fifteen corruption datasets (repairing the robustness on corrupted datasets)}) on both CIFAR-10 and Tiny-ImageNet. For CIFAR-10 dataset, an execution of training and repairing a model under one specific configuration costs about 12 hours on average (the maximum one is about 50 hours); while for Tiny-ImageNet dataset, an execution of training and repairing a model takes about 18 hours on average (the maximum one is about 64 hours). Overall, the total execution time of our experiments is more than 2 months.


\section{Experimental Results} 

In this section, we summarize the high-level results and findings for answering our research questions.

\begin{table*}[t]
    \centering
    \caption{Accuracy (\%) of 4 different DNNs (\ie, ResNet-18. ResNet-50, ResNet-101, and DenseNet-121) repaired on 2 dataset (\ie, CIFAR-10 and Tiny-ImageNet) by different repairing methods. }
    \resizebox{1\linewidth}{!}{
    \begin{tabular}{l|cccc|cccc}
    \toprule
        \multirow{2}{*}{\bf Baseline} & \multicolumn{4}{c|}{\bf CIFAR-10} & \multicolumn{4}{c}{\bf Tiny-ImageNet} \\
        & \bf ResNet-18 & \bf ResNet-50 & \bf ResNet-101 & \bf DenseNet-121 & \bf ResNet-18 & \bf ResNet-50 & \bf ResNet-101 & \bf DenseNet-121 \\
    \midrule
        \bf Original      & 85.00     & 85.17  & 85.72  & 87.97 & 45.15  & 46.27  & 46.14 & \cellcolor{tab_red}48.73 \\
    \midrule
        \bf MODE~\cite{ma2018mode}          & 85.13  & 85.26  & 86.19  & 88.28 & 45.75  & 45.93  & 45.87 & 47.69 \\
        \bf Apricot~\cite{zhang2019apricot}       & 86.80 & 88.95 & 89.74  & 89.93 & 46.30 & 46.85  & 45.90 & 45.27 \\
        \bf Arachne~\cite{sohn2019arachne}       & 85.38 & 87.95  & 89.37  & 91.25 & 46.73 & 47.37 & 46.75 & 46.95 \\
    \midrule
        \bf SENSEI~\cite{gao2020sensei}        & 85.20 & 86.25 & 88.73 & 89.73 & 45.82 & 46.92 & 46.38 & 45.38 \\
        \bf Few-Shot~\cite{ren2020fewshot}      & 86.28 & 86.35 & 88.28 & 88.57 & 45.82 & 46.92 & 45.87 & 45.26 \\
        \bf DeepRepair~\cite{yu2021deeprepair}    & 87.20 & 87.46 & 88.94 & 90.56 & 46.78  & 47.69 & \cellcolor{tab_red}46.94 & 46.97 \\
    \midrule
        \bf \method{} (ours)       & \cellcolor{tab_red}88.29 &  \cellcolor{tab_red}89.58 & \cellcolor{tab_red}90.38 & \cellcolor{tab_red}91.37 & \cellcolor{tab_red}47.35 & \cellcolor{tab_red}47.82 & 46.73 & 46.84 \\
    \bottomrule
    \end{tabular}
    }
    \label{tab:rq1}
\end{table*}

\begin{figure*}[t]
    \centering
    \includegraphics[width=0.9\linewidth]{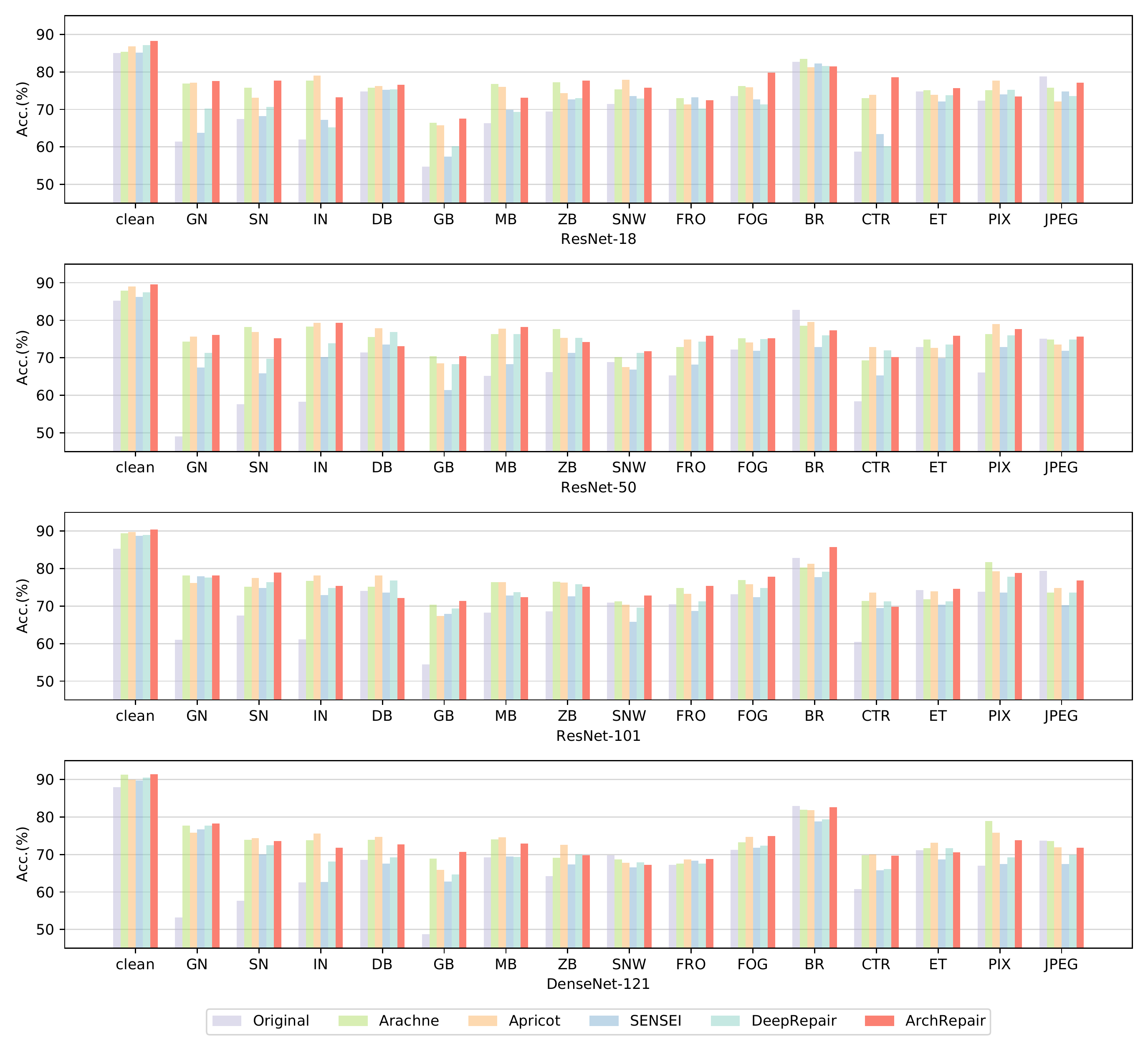} 
    \caption{Comparing the repairing methods on different DNNs (\ie, ResNet-18, ResNet-50, ResNet-101 and DenseNet-121) by contrasting the accuracy of repaired DNNs on CIFAR-10's testing dataset (\ie, $\mathcal{D}^\text{t}$) and corruption datasets (\ie, $\mathcal{D}^\text{c}$).} 
    \label{fig:rq1_cifar_bar}
\end{figure*}

\begin{figure*}[t]
    \centering
    \includegraphics[width=0.9\linewidth]{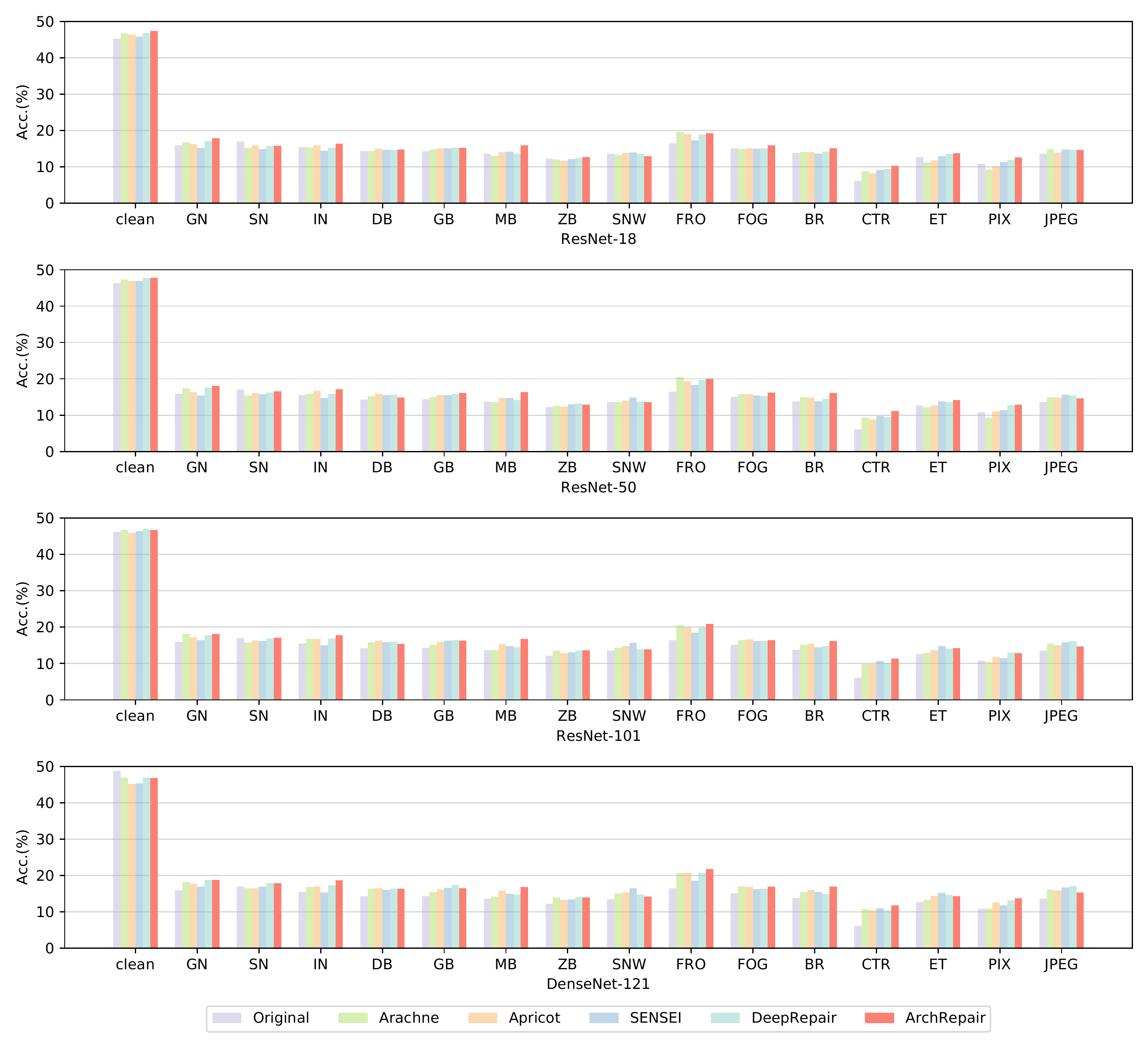} 
    \caption{Comparing the repairing methods on different DNNs (\ie, ResNet-18, ResNet-50, ResNet-101 and DenseNet-121) by contrasting the accuracy of repaired DNNs on Tiny-Imagenet's testing dataset (\ie, $\mathcal{D}^\text{t}$) and corruption datasets (\ie, $\mathcal{D}^\text{c}$).} 
    \label{fig:rq1_tiny_imagenet_bar}
\end{figure*}

\subsection{RQ1: Does \method{} outperform the state-of-the-arts (SOTA) DNN repair techniques?}\label{subsec:exp-rq1}

To answer RQ1, we train 4 DNNs (\ie, ResNet-18, ResNet-50, ResNet-101, and DenseNet-101) on CIFAR-10's and Tiny-ImageNet's training dataset (\ie, $\mathcal{D}^\text{t}$) and evaluate them on testing datasets (\ie, $\mathcal{D}^\text{v}$) respectively. To evaluate the performance of our method (\ie, \method{}), we apply six different SOTA methods as well as \method{} to repair these 4 DNNs. The evaluation results of repairing are summarized in \tableref{tab:rq1}. In general, \method{} exhibits significant advantages over all baseline methods on the 4 DNNs, demonstrating its effectiveness and generalization ability of the proposed method.
In particular, comparing with the state-of-the-art DNN repair methods (\ie, neuron-level repairing method Arachne~\cite{sohn2019arachne}, and network-level repairing method DeepRepair~\cite{yu2021deeprepair}), \method{} achieves much higher accuracy on all 4 DNNs on CIFAR-10 dataset. On the more challenging dataset, Tiny-ImageNet, \method{} still achieves much higher accuracy on 2 out of 4 DNNs. Note that on DenseNet-121, all the repairing methods failed to repair, \ie, didn't improve the performance comparing to the original network. One possible explanation is that the original DenseNet-121's performance has almost reached the upper-bound of the classification accuracy on Tiny-ImageNet (highest accuracy among 4 different DNNs), hence there might not be much room for improvement in terms of the accuracy.

Furthermore, to understand the influence of repairing on DNN's robustness, we evaluate the repaired DNNs' performance on corruption datasets (\ie, CIFAR-10-C~\cite{hendrycks2019robustness} and Tiny-ImageNet-C~\cite{hendrycks2019robustness}). The CIFAR-10-C and Tiny-ImageNet-C contain over 15 types of natural corruption datasets, and we show the results on CIFAR-10-C in \figref{fig:rq1_cifar_bar} and Tiny-ImageNet-C in \figref{fig:rq1_tiny_imagenet_bar}. Obviously in \figref{fig:rq1_cifar_bar}, \method{} achieves the highest accuracy on a majority of corruption datasets across three variants of ResNet (8/15, 9/15, and 7/15 on ResNet-18, ResNet-50, and ResNet-101, respectively) besides the best performance on clean dataset. 
Even on DenseNet-121, which is not a block-like DNN, \method{} also achieves promising performance compared with SOTA method Apricot~\cite{zhang2019apricot}. 
The performance of \method{} are also significant on Tiny-ImageNet-C. As we've mentioned before, Tiny-ImageNet is way more challenging. Nevertheless, \method{} still outperforms baselines in terms of the robustness on a majority of corruption datasets across three variants of ResNet (9/15, 9/15, and 7/15 on ResNet-18, ResNet-50, and ResNet-101, respectively) as well as the non-block-like DNN DenseNet-121 (8/15).
This fact confirms that \method{} doesn't harm the DNN's robustness, and on the contrary, it can even sometimes improve DNN's generalization ability towards classifying corrupted data.

\begin{tcolorbox}[size=title]
{\textbf{Answer to RQ1:} According to the experimental results on clean dataset, \method{} outperforms the SOTA repairing method on all 4 DNNs with different architetures (\ie, ResNet-18, ResNet-50, ResNet-101, and DenseNet-121). Moreover, the experimental results on corruption datasets also support that \method{} can repair a DNN without harming its robustness. }
\end{tcolorbox}

\begin{table*}
    \centering
    \caption{Accuracy (\%) of a deployed ResNet-18 repaired by different repairing method on 15 different corruption patterns.} 
    \resizebox{\linewidth}{!}{
    \begin{tabular}{c|l|cccccccccccccccc}
    \toprule
        \multicolumn{2}{c|}{\bf ResNet-18}
        & \bf Clean & \bf GN & \bf SN & \bf IN & \bf DB & \bf GB & \bf MB & \bf ZB & \bf SNW & \bf FRO & \bf FOG & \bf BR & \bf CTR & \bf ET & \bf PIX & \bf JPEG \\
    \midrule
        \multirow{6}{*}{\rotatebox{90}{CIFAR-10-C}} 
        & \bf Original      & 85.000 & 61.452 & 67.392 & 61.944 & 74.762 & 54.782 & 66.348 & 69.476 & 71.408 & 70.114 & 73.532 & 82.736 & 58.716 & 74.822 & 72.364 & 78.752 \\
        & \bf Apricot~\cite{zhang2019apricot}       & 86.644 & 76.930 & \cellcolor{tab_red}78.656 & 77.694 & 75.827 & 66.390 & \cellcolor{tab_red}76.810 & \cellcolor{tab_red}79.851 & 76.406 & 77.269 & 78.979 & \cellcolor{tab_red}89.254 & 74.390 & 75.112 & 75.350 & 75.810 \\
        & \bf Arachne~\cite{sohn2019arachne}       & 88.451 & 77.144 & 77.715 & \cellcolor{tab_red}78.976 & 76.546 & 65.815 & 75.963 & 77.712 & 77.862 & 77.224 & 79.200 & 86.913 & 75.792 & 73.876 & \cellcolor{tab_red}77.694 & 74.402 \\
        & \bf SENSEI~\cite{gao2020sensei}        & 86.525 & 68.762 & 70.471 & 73.345 & 76.842 & 60.244 & 71.229 & 73.297 & 73.732 & 73.814 & 76.975 & 83.006 & 64.861 & 72.814 & 75.833 & \cellcolor{tab_red}79.495 \\
        & \bf DeepRepair~\cite{yu2021deeprepair}    & 88.159 & 75.197 & 73.990 & 75.807 & 77.369 & 63.263 & 75.703 & 74.973 & 76.999 & 76.872 & 77.884 & 83.967 & 72.889 & 76.594 & 74.669 & 77.726 \\
        & \bf \method{} (ours)     & \cellcolor{tab_red}90.177 & \cellcolor{tab_red}77.546 & 77.689 & 73.237 & \cellcolor{tab_red}80.679 & \cellcolor{tab_red}67.523 & 75.998 & 77.697 & \cellcolor{tab_red}77.867 & \cellcolor{tab_red}80.677 & \cellcolor{tab_red}79.854 & 85.146 & \cellcolor{tab_red}79.026 & \cellcolor{tab_red}78.053 & 77.448 & 77.967 \\
    \midrule
        \multirow{6}{*}{\rotatebox{90}{Tiny-ImageNet-C}} 
		& \bf Original   & 45.150 & 15.912 & \cellcolor{tab_red}16.972 & 15.482 & 14.281 & 14.337 & 13.648 & 12.191 & 13.562 & 16.452 & 15.119 & 13.823 & 6.130 & 12.657 & 10.819 & 13.577 \\
		& \bf Apricot~\cite{zhang2019apricot}  & 46.732 & 16.703 & 15.270 & 15.339 & 14.266 & 14.762 & 13.047 & 11.959 & 13.319 & \cellcolor{tab_red}19.550 & 14.838 & 14.041 & 8.790 & 11.231 & 9.227 & \cellcolor{tab_red}14.825 \\
		& \bf Arachne~\cite{sohn2019arachne}   & 46.297 & 16.302 & 15.932 & 15.932 & \cellcolor{tab_red}14.938 & 15.152 & 14.119 & 11.695 & 13.805 & 18.986 & 15.106 & 14.123 & 8.253 & 11.831 & 10.145 & 13.918 \\
		& \bf SENSEI~\cite{gao2020sensei}      & 45.824 & 15.270 & 14.870 & 14.390 & 14.664 & 15.052 & 14.191 & 12.112 & \cellcolor{tab_red}13.917 & 17.250 & 14.943 & 13.602 & 9.117 & 12.902 & 11.277 & 14.772 \\
		& \bf DeepRepair~\cite{yu2021deeprepair}       & 46.780 & 17.032 & 15.673 & 15.277 & 14.669 & \cellcolor{tab_red}15.324 & 13.570 & 12.478 & 13.624 & 18.950 & 15.152 & 14.145 & 9.385 & 13.496 & 11.926 & 14.597 \\
		& \bf \method{{}} (ours)         & \cellcolor{tab_red}47.350 & \cellcolor{tab_red}17.820 & 15.779 & \cellcolor{tab_red}16.376 & 14.769 & 15.224 & \cellcolor{tab_red}15.967 & \cellcolor{tab_red}12.670 & 12.923 & 19.295 & \cellcolor{tab_red}15.915 & \cellcolor{tab_red}15.112 & \cellcolor{tab_red}10.337 & \cellcolor{tab_red}13.765 & \cellcolor{tab_red}12.553 & 14.624 \\
    \bottomrule
    \end{tabular}
    }
    \label{tab:rq2}
\end{table*}

\subsection{RQ2: Can \method{} fix DNN on a certain failure pattern without sacrificing robustness on clean data and other failure patterns?}\label{subsec:exp-rq2}

In \secref{subsec:exp-rq1}, our investigation results demonstrated that \method{} will not affect DNN's robustness when repairing on the clean dataset. Hence in this section, we continue to validate whether our method harms DNN's robustness when repairing a specific failure pattern.

We first verify the repairing capability of \method{}. We repair a deployed DNN (\ie, ResNet-18) on each of the corruption datasets from CIFAR-10-C and Tiny-ImageNet-C, and compare the performance with the other repairing methods, where the results are summarized in \tableref{tab:rq2}. Comparing the experimental results on the corruption dataset, we see that all repairing methods have the capability to repair the failure patterns, except shot noise (SN) on Tiny-ImageNet-C (all repairing methods fail to repair this corruption pattern). Among these repairing techniques, our method \method{} has the highest accuracy on 8 out of 15 the corruption datasets on CIFAR-10-C dataset, and 9 out of 15 the corruption datasets on Tiny-ImageNet-C, respectively, demonstrating that \method{} exhibits the advantages in repairing failure patterns. 

\begin{figure*}
    \centering
    \includegraphics[width=0.9\linewidth]{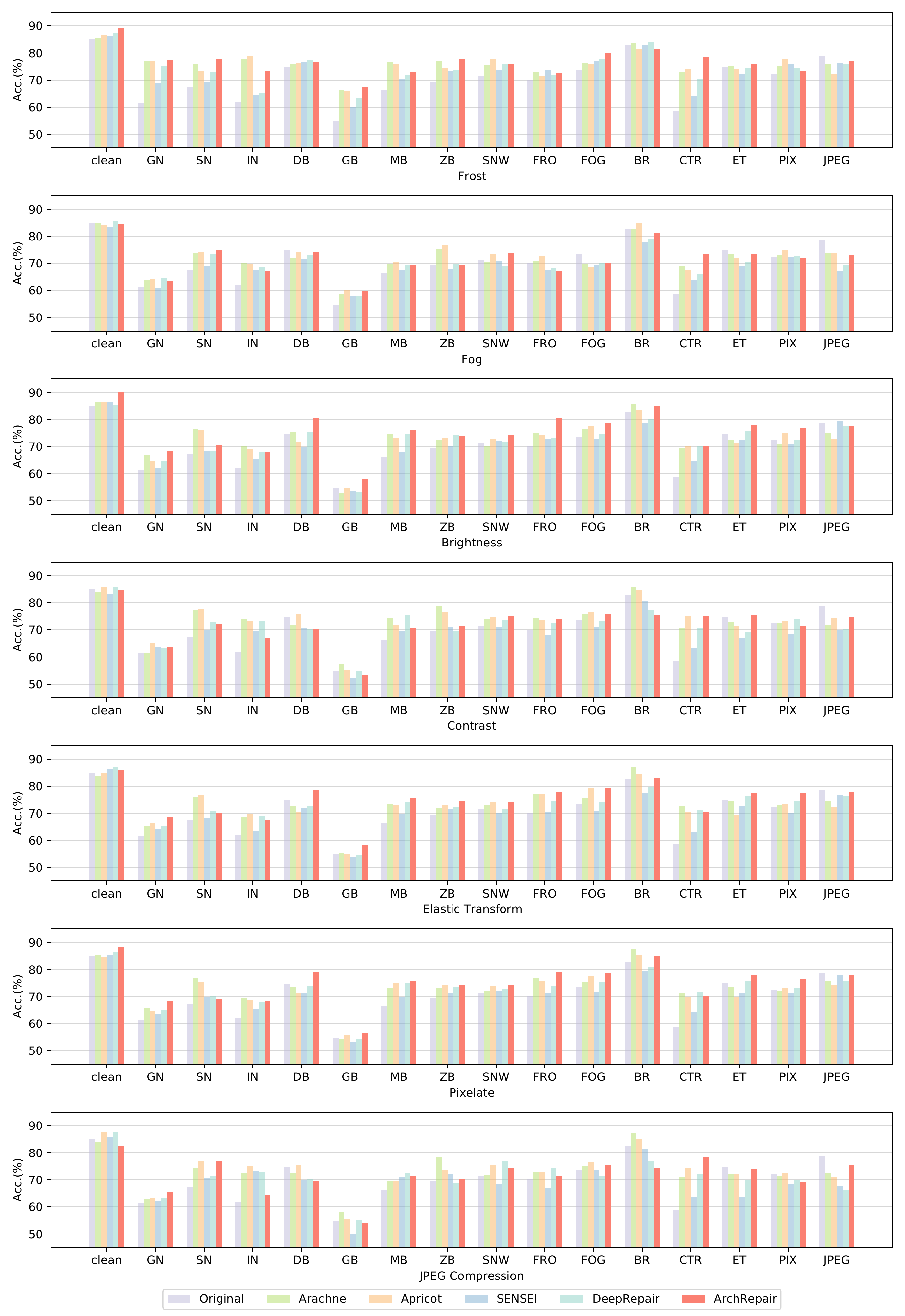}
    \caption{Comparing the effectiveness and robustness of repairing methods on ResNet-18 by repairing the DNNs on one of the CIFAR-10's corruption dataset $\mathcal{D}^\text{c}_\text{i}$ (CIFAR-10-C) and evaluating on the other corruption dataset $\{\mathcal{D}^\text{c}_\text{k} | \mathcal{D}^\text{c}_\text{k} \in \mathcal{D}^\text{c}, \text{k} \neq \text{i}\}$.} 
    \label{fig:rq2_cifar_bar}
\end{figure*}

\begin{figure*}
    \centering
    \includegraphics[width=0.9\linewidth]{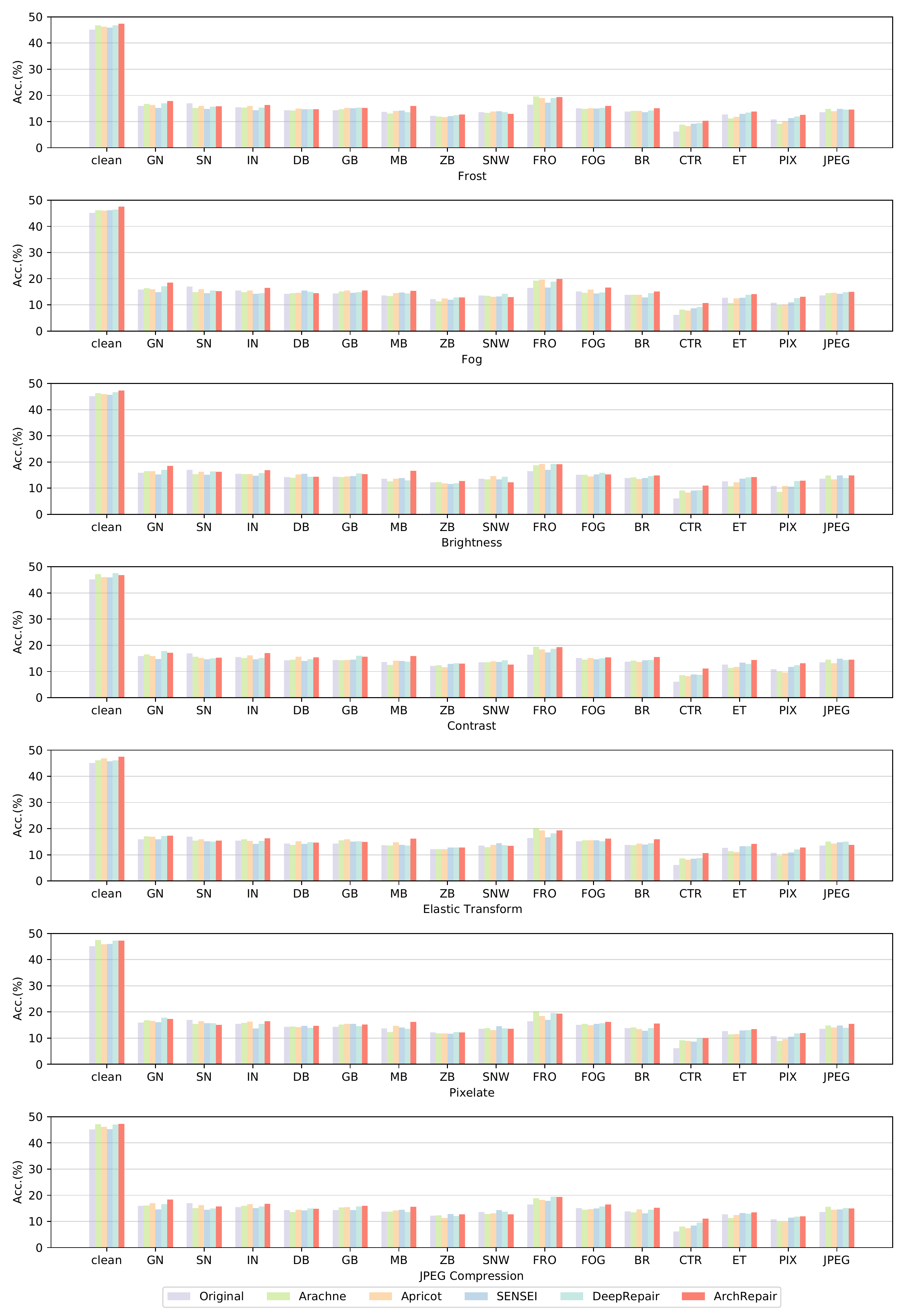}
    \caption{Comparing the effectiveness and robustness of repairing methods on ResNet-18 by repairing the DNNs on one of the Tiny-Imagenet's corruption dataset $\mathcal{D}^\text{c}_\text{i}$ (Tiny-ImageNet-C) and evaluating on the other corruption dataset $\{\mathcal{D}^\text{c}_\text{k} | \mathcal{D}^\text{c}_\text{k} \in \mathcal{D}^\text{c}, \text{k} \neq \text{i}\}$.} 
    \label{fig:rq2_tiny_imagenet_bar}
\end{figure*}

To validate whether our method has harmed DNN's robustness, we also evaluate the performance of repaired DNNs on the other corruption datasets. The evaluation results on CIFAR-10 and Tiny-ImageNet are shown in \figref{fig:rq2_cifar_bar} and \figref{fig:rq2_tiny_imagenet_bar}, respectively. Comparing the accuracy difference on CIFAR-10-C (see \figref{fig:rq2_cifar_bar}), we observe that the DNNs repaired by \method{} (\ie, the red bar) have higher accuracies on both clean and corruption datasets than the original DNN (\ie, the gray bar, which is lower than others in most of the cases), indicating that repairing method will not harm the DNN's robustness when having fixed certain corruption patterns. This is also verified by the results on Tiny-ImageNet-C (see \figref{fig:rq2_tiny_imagenet_bar}), where repairing on a certain corruption pattern will not affect the DNN's robustness on clean dataset and other corruption patterns, instead, it can even significantly enhance the robustness in some cases (\eg, when repairing on Fog corruption, the performance on other corruptions is also improved).

\begin{figure*}
    \centering
    \includegraphics[width=0.9\linewidth]{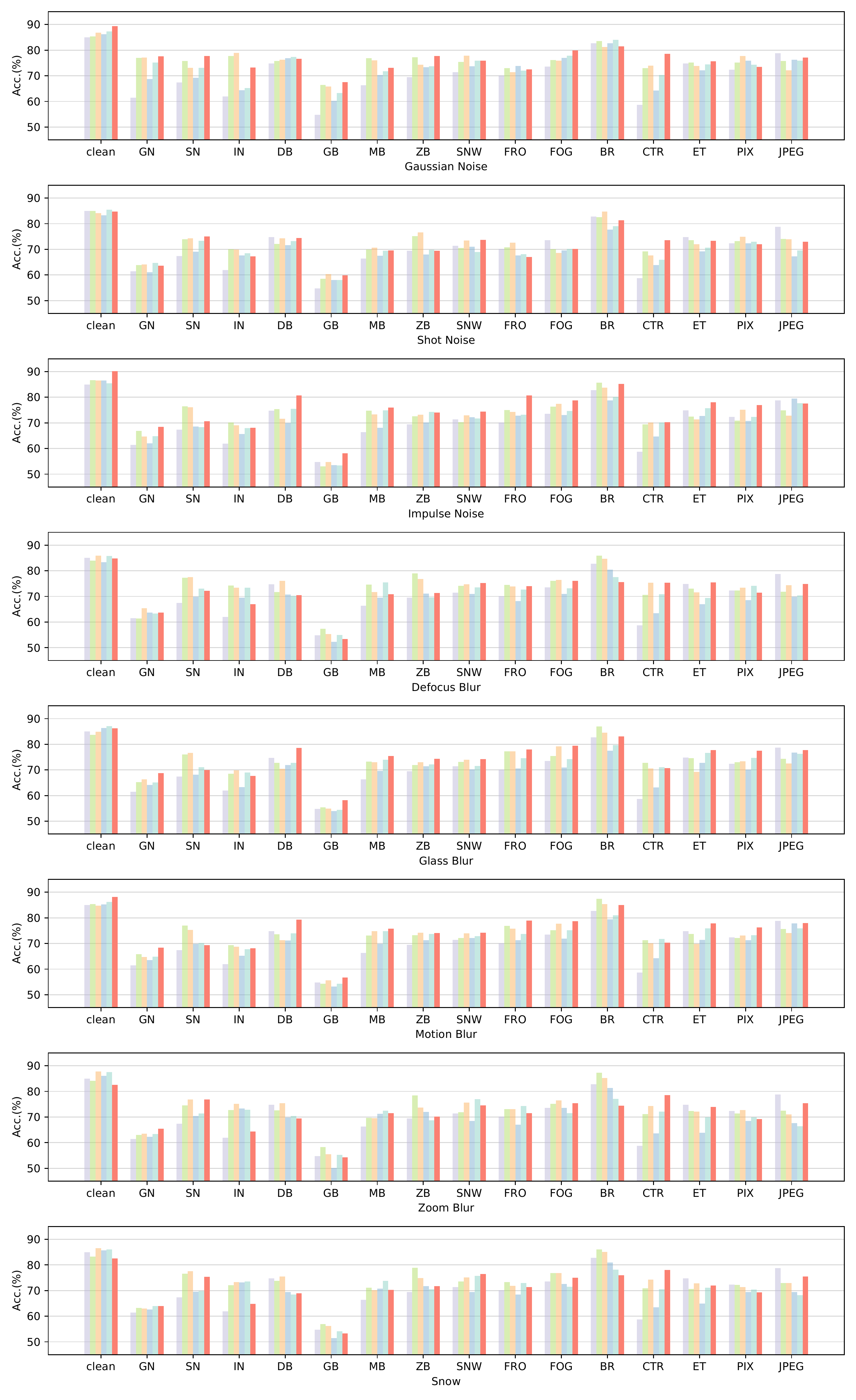}
\end{figure*}

\begin{figure*}
    \centering
    \includegraphics[width=0.9\linewidth]{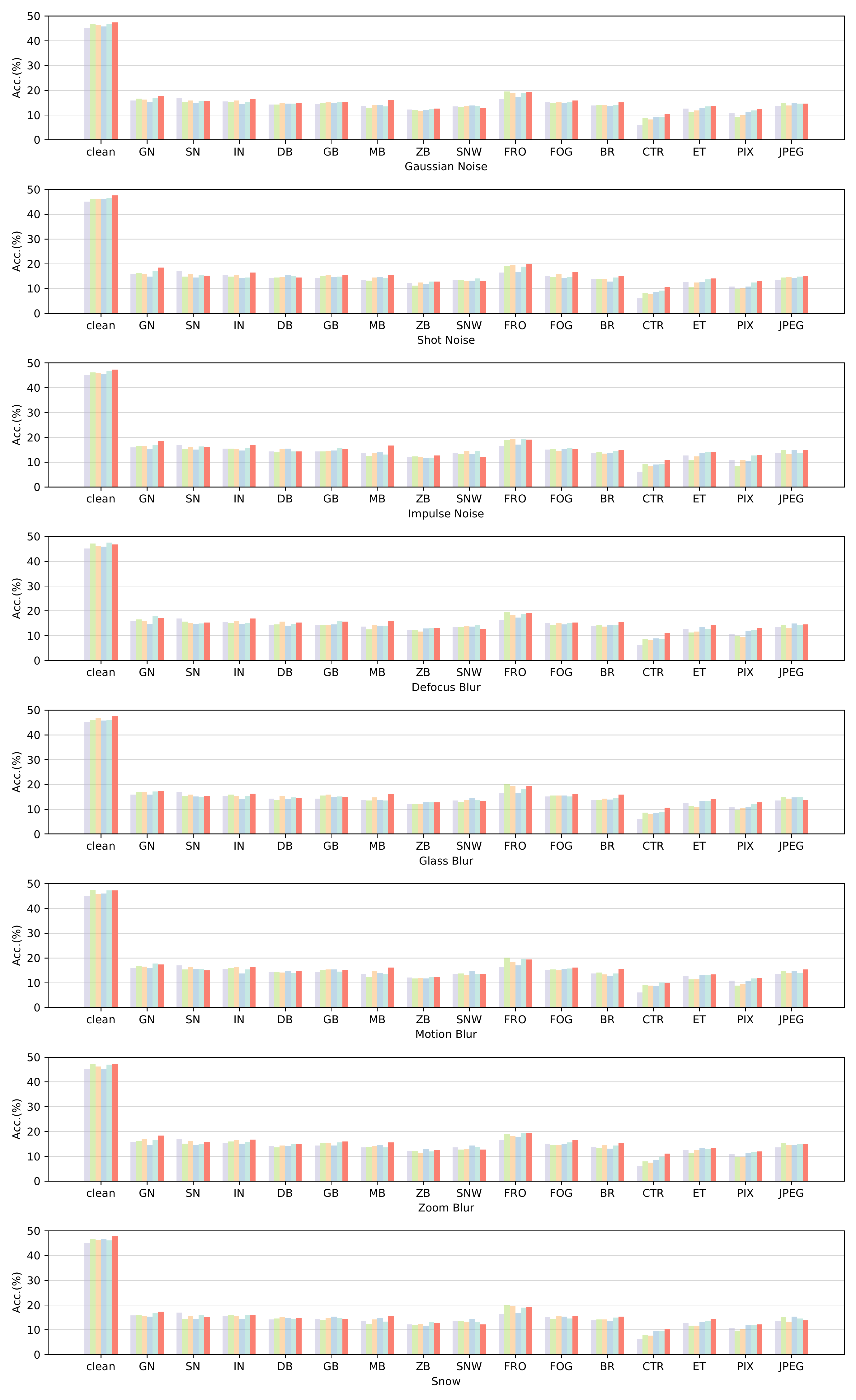}
\end{figure*}

\begin{tcolorbox}[size=title]
{\textbf{Answer to RQ2:} \method{} can successfully fix a certain corruption pattern on a deployed DNN (\ie, ResNet-18), outperforming the existing 4 DNN repair methods. In addition, \method{}'s repairing doesn't harm DNN's robustness on clean dataset and other failure patterns.}
\end{tcolorbox}

\begin{table*}[t]
    \centering
    \small
    \caption{Block suspiciousness $\mathcal{S}_\text{B}$ under 8 different thresholds $\epsilon_i$ and the accuracy of 2 DNNs (\ie, ResNet-18 and ResNet-50) repaired on 4 different blocks. Obviously repairing on the block with the highest block suspiciousness has the best performance.}
    \begin{subtable}[t]{\linewidth}
        \resizebox{\linewidth}{!}{
        \begin{tabular}{l|c|rrrrrrrr|c|rrrrrrrr}
        \toprule
            \multirow{3}{*}{} & \multicolumn{9}{c|}{\bf CIFAR-10} & \multicolumn{9}{c}{\bf Tiny-ImageNet} \\
            \cline{2-19}
            & \multirow{2}{*}{\bf Acc. (\%) on $\mathcal{D}^\text{v}$} & \multicolumn{8}{c|}{\bf Block Suspiciousness $\mathcal{S}_\text{B}$} & \multirow{2}{*}{\bf Acc. (\%) on $\mathcal{D}^\text{v}$} & \multicolumn{8}{c}{\bf Block Suspiciousness $\mathcal{S}_\text{B}$} \\
            &  & $\bm{\epsilon_{10}}$ & $\bm{\epsilon_{20}}$ & $\bm{\epsilon_{30}}$ & $\bm{\epsilon_{40}}$ & $\bm{\epsilon_{50}}$ & $\bm{\epsilon_{75}}$ & $\bm{\epsilon_{100}}$ & $\bm{\epsilon_{150}}$ &  & $\bm{\epsilon_{10}}$ & $\bm{\epsilon_{20}}$ & $\bm{\epsilon_{30}}$ & $\bm{\epsilon_{40}}$ & $\bm{\epsilon_{50}}$ & $\bm{\epsilon_{75}}$ & $\bm{\epsilon_{100}}$ & $\bm{\epsilon_{150}}$ \\
        \midrule
            \bf Block 1 & 85.374 &  0 &  3 &  6 &  8 &  8 & 18 & 22 & 40 & 46.11 & 1 & 1 & 4 & 4 & 4 & 12 & 23 & 41 \\
            \bf Block 2 & 86.377 &  0 &  0 &  0 &  1 &  1 &  2 &  5 & 16 & 46.29 & 0 & 1 & 2 & 2 & 2 & 6 & 9 & 16 \\
            \bf Block 3 & 85.090 &  0 &  1 &  3 &  9 & 17 & 19 & 26 & 47 & 47.13 & 0 & 0 & 0 & 1 & 4 & 5 & 9 & 16 \\
            \bf Block 4 & \cellcolor{tab_red}88.294 & \cellcolor{tab_red}10 & \cellcolor{tab_red}20 & \cellcolor{tab_red}21 & \cellcolor{tab_red}22 & \cellcolor{tab_red}24 & \cellcolor{tab_red}48 & \cellcolor{tab_red}48 & \cellcolor{tab_red}50 & \cellcolor{tab_red}47.35 & \cellcolor{tab_red}9 & \cellcolor{tab_red}18 & \cellcolor{tab_red}24 & \cellcolor{tab_red}33 & \cellcolor{tab_red}40 & \cellcolor{tab_red}52 & \cellcolor{tab_red}60 & \cellcolor{tab_red}79 \\
        \bottomrule
        \end{tabular}
        }
        \vspace{0.5pt}
        \caption{Block suspiciousness and repairing accuracy on ResNet-18}
    \end{subtable}\\
    \begin{subtable}[t]{\linewidth}
        \resizebox{\linewidth}{!}{
        \begin{tabular}{l|c|rrrrrrrr|c|rrrrrrrr}
        \toprule
            \multirow{3}{*}{} & \multicolumn{9}{c|}{\bf CIFAR-10} & \multicolumn{9}{c}{\bf Tiny-ImageNet} \\
            \cline{2-19}
            & \multirow{2}{*}{\bf Acc. (\%) on $\mathcal{D}^\text{v}$} & \multicolumn{8}{c|}{\bf Block Suspiciousness $\mathcal{S}_\text{B}$} & \multirow{2}{*}{\bf Acc. (\%) on $\mathcal{D}^\text{v}$} & \multicolumn{8}{c}{\bf Block Suspiciousness $\mathcal{S}_\text{B}$} \\
            &  & $\bm{\epsilon_{10}}$ & $\bm{\epsilon_{20}}$ & $\bm{\epsilon_{30}}$ & $\bm{\epsilon_{40}}$ & $\bm{\epsilon_{50}}$ & $\bm{\epsilon_{75}}$ & $\bm{\epsilon_{100}}$ & $\bm{\epsilon_{150}}$ &  & $\bm{\epsilon_{10}}$ & $\bm{\epsilon_{20}}$ & $\bm{\epsilon_{30}}$ & $\bm{\epsilon_{40}}$ & $\bm{\epsilon_{50}}$ & $\bm{\epsilon_{75}}$ & $\bm{\epsilon_{100}}$ & $\bm{\epsilon_{150}}$ \\
        \midrule
            \bf Block 1 & 82.115 &  1 &  2 &  2 &  4 &  4 & 7 & 7 & 7 & 45.83 & 0 & 0 & 0 & 0 & 0 & 0 & 0 & 0 \\
            \bf Block 2 & 84.313 &  1 &  1 &  6 &  8 &  8 &  10 &  10 & 15 & 46.55 & 0 & 0 & 0 & 0 & 0 & 0 & 0 & 3 \\
            \bf Block 3 & \cellcolor{tab_red}89.576 &  \cellcolor{tab_red}8 &  \cellcolor{tab_red}18 &  \cellcolor{tab_red}24 &  \cellcolor{tab_red}32 & \cellcolor{tab_red}42 & \cellcolor{tab_red}58 & \cellcolor{tab_red}86 & \cellcolor{tab_red}139 & \cellcolor{tab_red}47.82 & \cellcolor{tab_red}10 & \cellcolor{tab_red}20 & \cellcolor{tab_red}30 & \cellcolor{tab_red}40 & \cellcolor{tab_red}48 & \cellcolor{tab_red}67 & \cellcolor{tab_red}84 & \cellcolor{tab_red}119 \\
            \bf Block 4 & 87.254 & 0 & 0 & 0 & 0 & 0 & 0 & 0 & 0 & 46.27 & 0 & 0 & 0 & 0 & 0 & 1 & 2 & 3 \\
        \bottomrule
        \end{tabular}
        }
        \vspace{0.5pt}
        \caption{Block suspiciousness and repairing accuracy on ResNet-50}
    \end{subtable}

    \label{tab:rq3}
\end{table*}

\subsection{RQ3: Is our proposed localization effective in identifying vulnerable block candidates?}\label{subsec:exp-rq3}

To verify the effectiveness of our localization method, we conduct an experiment by applying the repairing method on all 4 blocks of ResNet-18~\&~ResNet-50, and comparing the accuracy on the clean datasets $\mathcal{D}^\text{v}$ of both CIFAR-10 and Tiny-ImageNet with their block suspiciousness $\mathcal{S}_\text{B}$ (\ie, the number of suspicious neurons in correspond block). We calculate the block suspiciousness under 8 different thresholds 
$\epsilon_i$~\footnote{$\epsilon_i$ indicates top-$i$ neurons with highest suspiciousness.} ($i\in\{10, 20, 30, 40, 50, 75, 100, 150\}$)
to evaluate how the threshold $\epsilon_i$ affects the block suspiciousness. The experimental results are summarized in \tableref{tab:rq3}.

As shown in \tableref{tab:rq3}, the block suspiciousness $\mathcal{S}_\text{B}$ of Block 4 in ResNet-18 and Block 3 in ResNet-50 are always the highest on both CIFAR-10 and Tiny-ImageNet datasets, no matter what value the threshold $\epsilon_i$ is. It matches the performance of repaired DNNs, where the DNN repaired on Block 4 in ResNet-18 and Block 3 in ResNet-50 has the highest accuracy, respectively. This demonstrates that our localization method can correctly locate the most vulnerable block. 

It's worth mentioning that for a simpler DNN architecture, \ie, ResNet-18, the vulnerable candidate block can be located more accurately when the threshold $\epsilon_i$ is small. As the threshold $\epsilon_i$ increases, the block suspiciousness $\mathcal{S}_\text{B}$ on other blocks becomes larger, making the localization method difficult to identify the vulnerable block. While for ResNet-50 (a relatively complex DNN), no matter what value the threshold $\epsilon_i$ is, the localization result are always significantly accurate (with a much higher suspiciousness $\mathcal{S}_\text{B}$ comparing with other blocks). 

\begin{tcolorbox}[size=title]
{\textbf{Answer to RQ3:} \method{} can always locate the most vulnerable block regardless the settings of threshold $\epsilon_i$ on different DNNs' architectures (\eg, ResNet-18 and ResNet-50).}
\end{tcolorbox}

\subsection{RQ4: How different components of \method{} impact its overall performance?}\label{subsec:exp-rq4}

To demonstrate the effectiveness of our \method{} and investigate how each component has contribute to its overall performance, we conduct an ablation study by repairing 4 pre-trained models (\ie, ResNet-18, ResNet-50, ResNet-101, and DenseNet-121) with two variants of our method on both CIFAR-10 and Tiny-ImageNet datasets. \tableref{tab:rq4} summarizes the evaluation results.
The first one performs \method{} on one single layer of the DNN, and we denote these variants as `Layer-lv' in \tableref{tab:rq4}. The second one is our full (complete) version that applies \method{} at the block level, we denote this variant as `Block-lv' in \tableref{tab:rq4}.

\begin{table*}[t]
    \centering
    \small
    \caption{Comparing the two variants of our methods on four DNNs by evaluating the accuracy of repaired DNN under testing dataset $\mathcal{D}^\text{t}$.}
    \resizebox{\linewidth}{!}{
    \begin{tabular}{c|cccc|cccc}
    \toprule
        \multirow{2}{*}{} & \multicolumn{4}{c|}{\bf CIFAR-10} & \multicolumn{4}{c}{\bf Tiny-ImageNet} \\
        & \bf ResNet-18 & \bf ResNet-50 & \bf ResNet-101 & \bf DenseNet-121 & \bf ResNet-18 & \bf ResNet-50 & \bf ResNet-101 & \bf DenseNet-121 \\
    \midrule
        \bf Original & 85.00 & 85.17 & 85.72 & 87.97 & 45.15 & 46.27 & 46.14 & \cellcolor{tab_red}48.73 \\
        \bf Layer-lv & 85.02 & 85.26 & 85.29 & 89.86 & 45.35 & 45.11 & 45.84 & 46.17 \\
        \bf Block-lv & \cellcolor{tab_red}88.29 & \cellcolor{tab_red}89.58 & \cellcolor{tab_red}90.38 & \cellcolor{tab_red}91.37 & \cellcolor{tab_red}47.35 & \cellcolor{tab_red}47.82 & \cellcolor{tab_red}46.73 & 46.84 \\
    \bottomrule
    \end{tabular}
    }
    \label{tab:rq4}
\end{table*}

Comparing with the original DNNs, the performance of `Layer-lv' is acceptable on CIFAR-10 dataset, as it slightly improves the behaviors on three DNNs (\ie, ResNet-18, ResNet-50, and DenseNet-121) and only decreases slightly on ResNet-101. The `Block-lv' achieves better performance on all of the four DNNs on CIFAR-10, and these results indicate that \method{}'s repairing capability is effective at both levels.
The performance on `Block-lv' is better than the `Layer-lv' on all the four DNNs on two different datasets, especially on the more challenging dataset Tiny-ImageNet, where `Layer-lv' only shows small improvement on ResNet-18 while `Block-lv' has significant improvement on all three variants of ResNet. This demonstrate that repairing on one specific layer cannot fully unleash \method{}'s potential while repairing on a block enables to take the advantage of all components of \method{}. Note that even though both `Block-lv' and `Layer-lv' fail to repair DenseNet-121 on Tiny-ImageNet (as well as all the SOTA baseline methods, see evaluation results in \tableref{tab:rq1}), `Block-lv' still performs better than `Layer-lv'.

\begin{tcolorbox}[size=title]
{\textbf{Answer to RQ4:} Block-level repairing is more effective than layer-level one towards fully releasing \method's repairing capability. In addition, adjusting the network's architecture and weights simultaneously is more effective than only adjusting the weights, especially for block-level repairing, demonstrating that jointly repairing the block architecture and weights is a promising research direction for DNN repair.}
\end{tcolorbox}

\subsection{Threat to validity}
The threats to the validity could be from the following aspects: 1) The selected dataset and the used model architectures could be a threat. To mitigate it, we selected the popular datasets as well as diverse architectures to evaluate our method. 
2) The selection of the corruption dataset could be biased, \ie, our method may not generalize well on other corruptions. We actually selected the 15 commonly used natural corruptions in the standard benchmarks of previous work~\cite{hendrycks2019robustness}.
3) A Further threat is from the implementation of our method as well as the usage of the existing baselines. To mitigate the threat,  we carefully follow the configuration as stated in the original papers or implementations, respectively. Moreover, our co-authors carefully test and review our code and the configuration of other tools.
Furthermore, to be comprehensive for better understanding the position of \method{}, we perform a large scale comparative study against 6 SOTA DNN repair techniques. The results confirm DNN repair could be even more promising and there are still opportunities ahead when going beyond focusing on repairing DNN weights only.

\section{Related Work}\label{sec:related}

\subsection{DNN Testing}

DNN testing is an important and relevant technique to DNN repair, aiming to detect potential buggy issues of a DNN.
Some recent work focus on testing criteria design. For example, DeepXplore~\cite{pei2017deepxplore} proposes the neuron coverage based on the number of activated neurons on given testing data, where the neuron coverage represents the adequacy of the testing data. 
Similarly, DeepGauge~\cite{ma2018deepgauge} proposes multi-granularity testing criteria, which are based on neural behaviors.
Different from previous work focusing on single neuron's behaviors, DeepCT~\cite{ma2019deepct} considers the interactions between the different neurons, and Kim \etal~\cite{kim2019guiding} propose the coverage criteria to measure the surprise of the inputs.
Some researchers~\cite{sekhon2019towards,ncmislead} also point out that the neuron coverage might fail if most of the neurons are activated by a few test cases, and more further research is still needed along this line.  

These testing criteria lay the foundation for testing generation techniques to detect defects in DNNs. DeepTest~\cite{tian2018deeptest} generates test cases based on the guidance of neuron coverage. TensorFuzz~\cite{odena2018tensorfuzz} proposes a distance-based coverage-guided fuzzing techniques to test DNNs. Similarly, DeepHunter~\cite{xie2019deephunter} proposes another coverage-guided testing technique by integrating the coverage criteria from DeepGauge. Readers can also see~\cite{ma2018deepmutation}. DeepStellar~\cite{du2019deepstellar} employs the coverage criteria and fuzzing technique for testing the recurrent neural network. More discussions on the progress of deep learning testing can be referred to the recent survey~\cite{dltestsurvey,arxiv18_sdle}.
Different from these testing techniques, our work mainly focuses on repairing DNNs and enhance their robustness and generalization ability, which can be considered as the downstream tasks of DNN testing.

\subsection{Fault Localization on Deep Neuron Network}
Fault localization aims to locate the root causing of software failures. Similar approaches have been widely studied for traditional software, which focus on developing faults identification methods such as spectral-based~\cite{jones2005tarantula,abreu_practical_2009,landsberg_evaluation_2015,landsberg_optimising_2018,naish_model_2011,zhang_theoretical_2017,perez_test-suite_2017}, model-based~\cite{birch_fast_2019,s._alves_method_2017}, slice-based~\cite{alves_fault-localization_2011}, and semantic fault localization~\cite{christakis_semantic_2019}.
Several works recently introduce fault localization on DNNs to find vulnerable neurons and repair their weights. Representative techniques include sensitivity-based fault localization~\cite{sohn2019arachne} and spectrum-based fault localization~\cite{eniser2019deepfault}. Eniser \etal~\cite{eniser2019deepfault} try to identify suspicious neurons responsible for unsatisfactory DNN performance, which is an early attempt to introduce fault localization technique on DNNs with promising results. 
However, these methods only consider a fixed DNN architecture and neuron-aware buggy behaviors, which is less flexible for real-world applications. 
Our work repairs DNN at a higher level (\ie, block level) by localizing the vulnerable block and jointly repairing the block architecture and weights, which is novel and havn't been investigated before.

\subsection{DNN Repair}
So far, there are several attempts for repairing DNN models. 
Inspired by software debugging, Ma \etal~\cite{ma2018mode} propose a novel model debugging technique for neural network models, which is denoted as MODE. MODE first performs state differential analysis on hidden layers to identify the faulty neurons that are responsible for the misclassification. Then, an input selection algorithm is used to select new input samples to retrain the faulty neurons.

Zhang \etal~\cite{zhang2019apricot} propose a weight-adjustment approach called Apricot to fix the DNN. Apricot first generates a set of reduced DNNs from the original model and trains them with a random subset of the original training dataset, respectively. For each failure example, Apricot separates reduced DNN models into two partitions, one successfully predicts the label and the other not, and takes the mean of the corresponding weight assignments of two partitions. After that, Apricot automatically adjusts the weight with these mean values.
Further, Sohn \etal~\cite{sohn2019arachne} propose a search-based repair technique for DNNs, called Arachne. Unlike other techniques, Arachne directly manipulates the neuron weights without retraining. Arachne first uses positive and negative input data to retain correct behavior and generate a patch, respectively. Then uses Particle Swarm Optimization (PSO) to search and locate faulty neurons, and uses the result of PSO candidate to update neurons’ weight, and further calculates fitness value based on the outcomes.

Recently, Gao \etal~\cite{gao2020sensei} have proposed a new algorithm called SENSEI, which uses guided test generation techniques to address the data augmentation problem for robust generalization of DNNs under natural environmental variations. Firstly, SENSEI uses a genetic search on a space of the natural environmental variants of each training input data to identify the worst variant for augmentation on each epoch. Besides, SENSEI uses a heuristic technique named selective augmentation, which allows skipping augmentation in certain epochs based on an analysis of the DNN’s current robustness.
Another recent attempt for DNN repair is DeepRepair~\cite{yu2021deeprepair}, a method to repair the DNN on the image classification task. DeepRepair uses a style-guided data augmentation for DNN repairing to introduce the unknown failure patterns into the training data to retrain the model and applies clustering-based failure data generation to improve the effectiveness of data augmentation. 

Our repairing method is orthogonal to data-augmentation based methods such as SENSEI~\cite{gao2020sensei} and DeepRepair~\cite{yu2021deeprepair}, where we focus on repairing DNN from the architecture and weight perspective. Our method also goes one step further beyond the weight level (\eg, MODE~\cite{ma2018mode}, Apricot~\cite{zhang2019apricot}, and Arachne~\cite{sohn2019arachne}), and considers at a higher granularity by jointly repairing architecture and weights at block level, which is demonstrated to be a promising direction for DNN repairing.

\subsection{Neural Architecture Search}
Neural architecture search (NAS) could be another relevant line of our work, aiming to automatically design an architecture instead of handcrafting one.
Typical NAS includes evolution-based~\cite{real_regularized_2019,xie_genetic_2017}, and reinforcement-learning-based~\cite{baker_designing_2017} methods. 
However, the resources RL or evolution-based methods leveraged are often very expensive and still unaffordable in practice. 
More recently, DARTS~\cite{liu2018darts} relaxes the search space to make it continuous so that the search processes can be performed based on the gradient. %
Differentiable NAS approaches can significantly reduce the computational cost.
Our search method is based on PC-DARTS~\cite{xu_pc-darts:_2020}, a stability improved variant of DARTS by introducing a partially connected mechanism. 

The purpose of repairing and NAS is very different. The former intends to fix the buggy behaviors that follow some patterns with generalization capability, while NAS is to design general architecture automatically for better performance (\eg, energy efficiency).
%
In this paper, we formulate the block-level joint architecture and weight repairing as a NAS problem, which demonstrates the possibilities and chances for DNN repair along this direction.

\section{Conclusion}\label{sec:concl}

In this work, we have proposed \method{}, an architecture-oriented DNN repair at block level, which offers a good trade-off between repaired network accuracy and time consumption, compared to neuron-level, layer-level, and network-level (data augmentation) repairing. To achieve this, two key problems are identified and solved sequentially, \ie, \emph{block localization}, and \emph{joint architecture and weights repairing}. 
By jointly repairing both architecture and weights on the candidate block for repairing, \textbf{\emph{ArchRepair}} is able to achieve better repairing performance compared with 6 SOTA techniques.
Our extensive evaluation have also demonstrated that our method could not only enhance the accuracy but also the robustness across various corruption patterns while being cost-effective.
To the best of our knowledge, this is the very first attempt about DNN repair by considering adjusting both the architecture and weights at the `block-level'. Our research also initiates a promising direction for further DNN repair research, towards addressing the current urgent industrial demands for reliable and trustworthy DNN deployment in diverse real-world environments.

%
%
%
\bibliographystyle{splncs04}
\bibliography{ref_llncs}

\end{document}